\def\year{2020}\relax
\documentclass[letterpaper]{article}
\usepackage{template/aaai20}
\usepackage{times}
\usepackage{helvet}
\usepackage{courier}
\usepackage{url}
\usepackage{graphicx}
\usepackage[utf8]{inputenc} 
\usepackage[T1]{fontenc}    
\usepackage{url}            
\usepackage{booktabs}       
\usepackage{amsfonts}       
\usepackage{nicefrac}       
\usepackage{microtype}      
\usepackage{amsmath}
\usepackage{bm}  
\usepackage{balance} 
\usepackage{xcolor}
\usepackage{float} 
\usepackage{subcaption}
\usepackage{amsmath}
\usepackage{cleveref}
\usepackage{xr}
\usepackage{etoolbox}
\usepackage[ruled]{algorithm2e}

\DeclareMathOperator*{\argmin}{arg\,min}

\def\citationlevel#1{\def\citelevel{#1}}
\def\citeif#1{\expandafter\citeiff#1\relax}
\def\citeiff#1,#2\relax{\ifnum\numexpr#1-1<\citelevel\relax\cite{#2}\else\unskip\fi}
\citationlevel{1}

\usepackage{mathtools}

\def\citationlevel#1{\def\citelevel{#1}}
\def\citeif#1{\expandafter\citeiff#1\relax}
\def\citeiff#1,#2\relax{\ifnum\numexpr#1-1<\citelevel\relax\cite{#2}\else\unskip\fi}
\citationlevel{1}

\usepackage{mathtools}
\newcommand{\bb}{\mathbf{b}}

  \newcommand{\bgg}{\mathbf{g}}

\newcommand{\xx}{\mathbf{x}}
\newcommand{\yy}{\mathbf{y}}
\newcommand{\zz}{\mathbf{z}}

\newcommand{\bAA}{\mathbf{A}}

\newcommand{\CC}{\mathbf{C}}

\newcommand{\HH}{\mathbf{H}}
\newcommand{\II}{\mathbf{I}}

\newcommand{\KK}{\mathbf{K}}

\newcommand{\PP}{\mathbf{P}}

\newcommand{\TT}{\mathbf{T}}
\newcommand{\UU}{\mathbf{U}}

\newcommand{\WW}{\mathbf{W}}
\newcommand{\XX}{\mathbf{X}}
\newcommand{\YY}{\mathbf{Y}}
\newcommand{\ZZ}{\mathbf{Z}}

\newcommand{\LLambda}{\bm{\Lambda}}

\newcommand{\Reals}{\mathbb{R}}	

\DeclarePairedDelimiter\floor{\lfloor}{\rfloor}
    
\definecolor{darkgreen}{rgb}{0.1, 0.8, 0.0}

\newcommand\JP[1]{\textcolor{darkgreen}{#1}}

\newcommand\KM[1]{\textcolor{blue}{#1}}
\newcommand{\KMcheck}[1]{\textcolor{orange}{[#1]}}

\newcommand{\PRETRAINED}{HSIC-trained} 
\newcommand{\PRETRAINING}{HSIC-training}
\newcommand{\UNFORMATTED}{unformatted-training}
\newcommand{\UNFORMATTEDC}{Unformatted training}
\newcommand{\UNFORMATTEDED}{unformatted-trained}
\newcommand{\POSTTRAINED}{format-trained}
\newcommand{\POSTTRAINEDING}{format-training}
\newcommand{\POSTTRAINEDC}{Format-trained}
\newcommand{\POSTTRAINING}{format-training}
\newcommand{\COMBINEDNETC}{Multiple-scale network}
\newcommand{\COMBINEDNET}{multiple-scale network}
\newcommand{\HSIC}{\text{HSIC}}
\newcommand{\nHSIC}{\text{nHSIC}}

\DeclareMathOperator{\tr}{tr}
\newcommand{\Fig}{Fig.}
\newcommand{\Figs}{Figs.}

\title{The HSIC Bottleneck: Deep Learning without Back-Propagation}

\author{
Wan-Duo Kurt Ma\hspace{0.2in}
J.P. Lewis\hspace{0.2in}
W. Bastiaan Kleijn\\
\Large{Victoria University}\\
mawand@ecs.vuw.ac.nz, jplewis@google.com, bastiaan.kleijn@ecs.vuw.ac.nz
}

\begin{document}

\maketitle

\begin{abstract}
We introduce the HSIC (Hilbert-Schmidt independence criterion) bottleneck for training deep neural networks. The HSIC bottleneck is an alternative to the conventional cross-entropy loss and backpropagation that has a number of distinct advantages. It mitigates exploding and vanishing gradients, resulting in the ability to learn very deep networks without skip connections. There is no requirement for symmetric feedback or update locking. We find that the HSIC bottleneck provides performance on MNIST/FashionMNIST/CIFAR10 classification \emph{comparable to backpropagation with a cross-entropy target}, even when the system is not encouraged to make the output resemble the classification labels. Appending a single layer trained with SGD (without backpropagation) to reformat the information further improves performance.  
\end{abstract}

\section{Introduction}

Deep learning has brought a new level of performance to an increasingly wide range of tasks.  In practice however, the stochastic gradient descent (SGD) algorithm (and its variants) and the associated error back-propagation algorithm underlying deep learning are time consuming, have problems of vanishing and exploding gradients, require sequential computation across layers and update locking, and typically require the  exploration of learning rates and other hyperparameters. At the same time, backpropagation is generally regarded as being not biologically plausible.
These considerations are driving research into both theoretical and practical alternatives. 

We propose a deep network training method that does not use the cross-entropy loss or backpropagation. An alternate information-theoretic motivation of training a classifier can be found in terms of the Fano inequality and mutual information. In our context, Fano's inequality indicates that the probability of classification error depends on the conditional entropy $H(Y|X)$, with $Y$ being the label and $X$ some representation of the input. Additionally, the mutual information $I(X,Y)$ can be written in the form $I(X,Y) = H(Y) - H(Y|X)$. Note that the entropy of the labels is constant with respect to the network weights. When the probability of mis-classification is low, $H(Y|X)$ is also low, and the mutual information is high. Mutual information thus provides a training objective that directly involves the representation $X$, unlike cross entropy, which involves $X$ only through backpropagation.

As is the case with cross entropy, an algorithm trained exclusively using mutual information is vulnerable to overfitting. To address this we train the network using an approximation of the information bottleneck \cite{Tishby99theinformation}. Due to the practical difficulties of calculating the mutual information among the random variables, we adopt a non-parametric kernel-based method, the Hilbert-Schmidt independence criterion (HSIC), to characterize the statistical (in)dependence of different layers. That is, for each network layer we maximize HSIC between the layer activation and the desired output and minimize HSIC between that layer activation and the input.  In some cases, the final-layer representations resulting from this \textit{HSIC bottleneck} training are nearly one-hot and can be \emph{directly used for classification} after identifying a single fixed permutation.  Alternately, following HSIC bottleneck training, we append a single classification layer that is trained with SGD (without backpropagation), consistently obtaining performance competitive with an architecturally equivalent network trained with backpropagation. We provide an informal discussion of the relation between HSIC and mutual information in the Appendix.

Our work joins an increasing body of recent research that explores deep learning fundamentals from an information theoretical perspective (\cite{ShwartzZivT17,michael2018on,Tishby99theinformation,Ishmael18} and others). Our contributions are as follows: 

We demonstrate that it is possible to train deep classification networks using an information bottleneck principle, without backpropagation, and obtain results competitive with the standard backpropagation optimization of the cross-entropy objective.
The HSIC bottleneck approach\footnote{Our code is available at 
\url{https://github.com/choasma/HSIC-Bottleneck}}
mitigates the issue of vanishing or exploding gradients in backpropagation. As the HSIC-bottleneck operates directly on continuous random variables, it is more attractive than conventional information bottleneck approaches based on binning. Because the network training is explicitly based on an information bottleneck principle, it addresses overfitting by design. It further addresses 
the weight transport and update locking problems of backpropagation.

The experimental results demonstrate training several simple textbook architectures using the HSIC bottleneck, together with the results of the otherwise identical architecture trained with backpropagation. However, note that the backpropagation results require an additional final layer and softmax that is not required nor used in the ``unformatted'' HSIC-training case.
While the results provide a fair comparison, there was no exploration of alternative architectures, regularization, or data augmentation, and relatively little effort towards finding optimal hyperparameters. Thus neither the backpropagation baseline nor our method give state-of-the-art results, and further improvements are likely possible.

\textbf{Notation.} Upper case (e.g.,~$X,Y$) denotes random variables. Bold denotes vectors of observations (lower case, e.g.,~$\xx,\yy$) 
or matrices (e.g.,~$\HH, \KK$). Hilbert spaces are denoted with calligraphic font (e.g.,~$\mathcal{G}, \mathcal{H}$).

\section{Background and Related Work} \label{s:related}
Although SGD using the backpropagation algorithm \cite{Werbos90} is the predominant approach to optimizing deep neural nets, other approaches have been considered \cite{balduzzi15kickback,Lillicrap2016,Theodore2018,KohanErrorforward18,Choromanska19}. Kickback \cite{balduzzi15kickback} follows the local gradient using a direction obtained from the global single-class error.
Feedback alignment \cite{Lillicrap2016} shows that deep neural networks can be trained using random feedback connections. The alternating minimization \cite{Choromanska19}, a coordinate descent-like approach, breaks the nested objective into a collection of subproblems by introducing auxiliary variables, thereby allowing layer-parallel updates. 

Information theory \cite{Cover2006} underlies much research on learning theory  \cite{Ishmael18,Kwak02,brakel2018learning} as well as thinking in neuroscience \cite{BaddeleyInfoTheoryBrain}. 
The information bottleneck (IB) principle \cite{Tishby99theinformation} generalizes the notion of minimal sufficient statistics, expressing a tradeoff in the hidden representation between the information needed for predicting the output, and the information retained about the input.
The IB objective is 
\begin{equation} \label{eq:ib}
\min_{p_{T_i|X}}  I(X; T_i) - \beta I(T_i; Y),
\end{equation}
where $X$, $Y$ are the input and label random variable respectively, and $T_i$ represents the hidden representation at layer $i$. 
Intuitively, the IB principal preserves the information of the hidden representations about the label while compressing  information about the input data.

The IB principle has been employed both to explore deep learning dynamics and as a training objective in a growing body of recent work \cite{ShwartzZivT17,michael2018on,goldfeld18,alemiFischerDeep,Wu2018,kolchinskyNIB,banerjeeDB,amjadGeiger18} and others.
 
In practice, the IB is hard to compute for several reasons. If the network inputs are regarded as continuous, the mutual information $I(X,T_i)$ is infinite unless noise is added to the network. Many algorithms are based on binning, which 
suffers from the curse of dimensionality
and yields different results with different choices of bin size. The distinction between discrete and continuous data, and between discrete and differential entropy, presents additional considerations \cite{michael2018on,goldfeld18}. These issues are clearly surveyed in \cite{amjadGeiger18}. In the case of continuous variables and a deterministic network, the true MI is infinite, while discrete data results in a piecewise-constant MI that is also unsuited for optimization.  Existing approaches to applying IB to DNN learning have resorted to MI approximations such as adding noise and computing a bound rather than the actual quantity. However tighter bounds may not be better \cite{tschannenMI19}. It has been argued that the IB has inherent problems when applied to deep neural nets, and that current results may reflect the approximations to MI and inductive biases of the networks more than the true underlying mutual information \cite{amjadGeiger18,tschannenMI19}.

In this paper, we replace the mutual information terms in the information bottleneck objective with HSIC. 
In contrast to mutual information based estimates, HSIC provides a robust computation with a 
time complexity $O(m^2)$ where $m$ is the number of data points.\footnote{In our context $m$ is the  minibatch size.} HSIC \cite{Gretton2005} is the Hilbert-Schmidt norm of the cross-covariance operator between the distributions in Reproducing Kernel Hilbert Space (RKHS). The formulation of HSIC is: 

\begin{equation} \label{eq:hsic_1}
\begin{split}
\HSIC(\mathbb{P}_{XY}, \mathcal{H}, &\mathcal{G}) = \left \| C_{XY} \right \|^{2} \\
&= \mathbb{E}_{XYX'Y'}[k_{X}(X,X')k_{Y'}(Y,Y')] \\
&+ \mathbb{E}_{XX'}[k_{X}(X,X')]\mathbb{E}_{Y'}[k_{Y}(Y,Y')] \\
&- 2\mathbb{E}_{XY}[\mathbb{E}_{X'}[k_{X}(X,X')]\mathbb{E}_{Y'}[k_{Y}(Y,Y')]],
 \end{split}
\end{equation}
where $k_{X}$ and $k_{Y}$ are kernel functions, $\mathcal{H}$ and $\mathcal{G}$ are the Hilbert spaces, and $\mathbb{E}_{XY}$ is the expectation over $X$ and $Y$.

Let $\mathcal{D} := \{(\mathbf{x}_1, \mathbf{y}_1), \cdots, (\mathbf{x}_m, \mathbf{y}_m)\}$ contain 
$m$ i.i.d. samples drawn from $\mathbb{P}_{XY}$, where $\mathbf{x}_i \in \mathbb{R}^{d_x}$ and $\mathbf{y}_i \in \mathbb{R}^{d_y}$. Then \eqref{eq:hsic_1} leads to the following empirical expression \cite{Gretton2005}:
\begin{equation} \label{eq:hsic}
\HSIC(\mathcal{D}, \mathcal{H}, \mathcal{G}) = (m-1)^{-2}\tr(\KK_X \HH \KK_Y \HH)
\end{equation}
where $\KK_X \in \mathbb{R}^{m \times m}$ and $\KK_Y \in \mathbb{R}^{m \times m}$ have entries ${\KK_X}_{ij} = k(\xx_i, \xx_j)$ and ${\KK_Y}_{ij} = k(\yy_i, \yy_j)$, and $\HH \in R^{m \times m}$ is the centering matrix $\HH = \II_m - \frac{1}{m} \textbf{1}_m\textbf{1}_m^T$. 

With an appropriate kernel choice such as the Gaussian 
 $k(\xx,\yy) \sim \exp( - \frac{1}{2} \| \xx - \yy \|^2 / \sigma^2) $, 
HSIC is zero if and only if the random variables  $X$ and $Y$ are independent, $P_{XY} = P_{X}P_{Y}$  \cite{sriperumbudur2010relation}.
An intuition for the HSIC approach is provided by the fact that the series expansion of the exponential contains a weighted sum of all moments of the data, and (under reasonable conditions) two distributions are equal if and only if their moments are identical.
Considering the expression \eqref{eq:hsic}, the i'th of component of \eqref{eq:hsic} is 
\newcommand{\tk}{\bar{k}}
\newcommand{\tl}{\bar{l}}
\begin{equation}
    \langle {k_{X}}_{i,1}, \, {k_{X}}_{i,2}, \, \cdots , {k_{X}}_{i,n} \rangle
    \cdot
    \langle {k_{Y}}_{1,i}, \, {k_{Y}}_{2,i}, \, \cdots , {k_{Y}}_{n,i} \rangle
\label{eq:hsicdotprod}
\end{equation}
where ${k_X}_{i,j} \equiv (\KK_X\HH)_{i,j}$ and similarly ${k_Y}_{i,j} \equiv (\KK_Y\HH)_{i,j}$.
This inner product will be large when the relation between each point $i$ of $X$ and all other points of $X$ is similar to the relation between the corresponding point $i$ of $Y$ and all other points of $Y$, summed over all $i$, and where similarity is measured through the kernel $k(x_i,x_j)$ that (appropriately chosen) captures all statistical moments of the data.

In our experiments we use the normalized-HSIC (\nHSIC) formulation based on the normalized cross-covariance operator \cite{Fukumizu08,Blaschko2008b}, 
given by:
\begin{equation} \label{eq:nhsic}
    \nHSIC(\mathcal{D}, \mathcal{H}, \mathcal{G}) = \tr(\widetilde{\KK}_X\widetilde{\KK}_Y)
\end{equation}
where $\widetilde{\KK}_X=\overline{\KK}_X\left(\overline{\KK}_X+\epsilon m \II_m\right)^{-1}$ and $\widetilde{\KK}_Y=\overline{\KK}_Y\left(\overline{\KK}_Y+\epsilon m \II_m\right)^{-1}$.  $\overline{\KK}_X$ and $\overline{\KK}_Y$ denote centered kernel matrices, and $\epsilon$ is a small constant.

Unlike mutual information, HSIC does not have an interpretation in terms of information theoretic quantities (bits or nats). On the other hand, HSIC does not require density estimation and is simple and reliable to compute. Moreover, kernel distribution embedding approaches such as HSIC can also be resistant to outliers, as can be seen by considering the effect of outliers under the Gaussian kernel. The empirical estimate converges to the population HSIC value at the rate $1/\sqrt{n}$ independent of the dimensionality of the data \cite{Gretton2005}, meaning that it partially circumvents the curse of dimensionality. 

HSIC is widely used as a dependency measurement, including in the deep learning literature. For example, \cite{Wu2018} investigated the generalization properties of autoencoders using HSIC, while \cite{Romain18} uses HSIC to restrict the latent space search to constrain the aggregate variational posterior. \cite{Vepakomma19} use distance correlation (an alternate formulation of HSIC) to remove unnecessary private information from medical training data.

While in principle HSIC can discover arbitrary dependencies between variables, in practice and with finite data the choice of the $\sigma$ parameter in the HSIC kernel emphasizes relationships at some scales more than others. Intuitively, two data points $x,y$ are not well distinguished when their difference is sufficiently small or large, such that they lie on the small-slope portions of the Gaussian.
This is typically handled by choosing the kernel $\sigma$ based on median distances among the data \cite{Dino12,Sugiyama12}, or by a hyperparameter search\ifdef{\LONGVERSION}{(such as grid-search \citeif{2,Hinton2012} or random-search \citeif{Bergstra12}).}{.}

In our case, the data ``points'' are minibatches of activations from different network layers, and thus have different dimensionality. This suggests that different $\sigma$ values should be used in each kernel. Using the observation that the expected squared distance between random points scales with dimension, these multiple hyperparameters can be approximated by scaling a single $\sigma$ with the dimensionality $d$ of points $\xx$ and $\yy$ as: $k(\mathbf{x},\mathbf{y}) \sim \exp( - \frac{1}{2} \| \xx - \yy \|^2 / (\sigma^2 d))$ (used in all our experiments).

\section{Proposed Method} \label{s:proposed}


In this section we introduce the proposed \PRETRAINED\ network. Training a deep network without backpropagation using the HSIC-bottleneck objective will be generally termed \emph{HSIC-bottleneck training} or \emph{\PRETRAINING}. The output of the bottleneck-trained network contains the information necessary for classification, but not necessarily in the right form. We evaluate two specific approaches to produce classifications from the \textit{HSIC-bottleneck trained} network. First, if the outputs are one-hot, they can simply be permuted to align with the training labels. This is termed \emph{\UNFORMATTED.} In the second scheme, we append a single layer and softmax output to the frozen \UNFORMATTEDED network, and train the appended layer using SGD without backpropagation. Since this step is ``reformatting'' the information, this step is termed \textit{\POSTTRAINING.}
Note that the networks trained with backpropagation contain this softmax layer in every case.

\begin{figure}[ht] 
  \centering
  \hspace{1mm}
  \begin{subfigure}[b]{\columnwidth} 
    \centering
    \includegraphics[width=0.8\columnwidth]{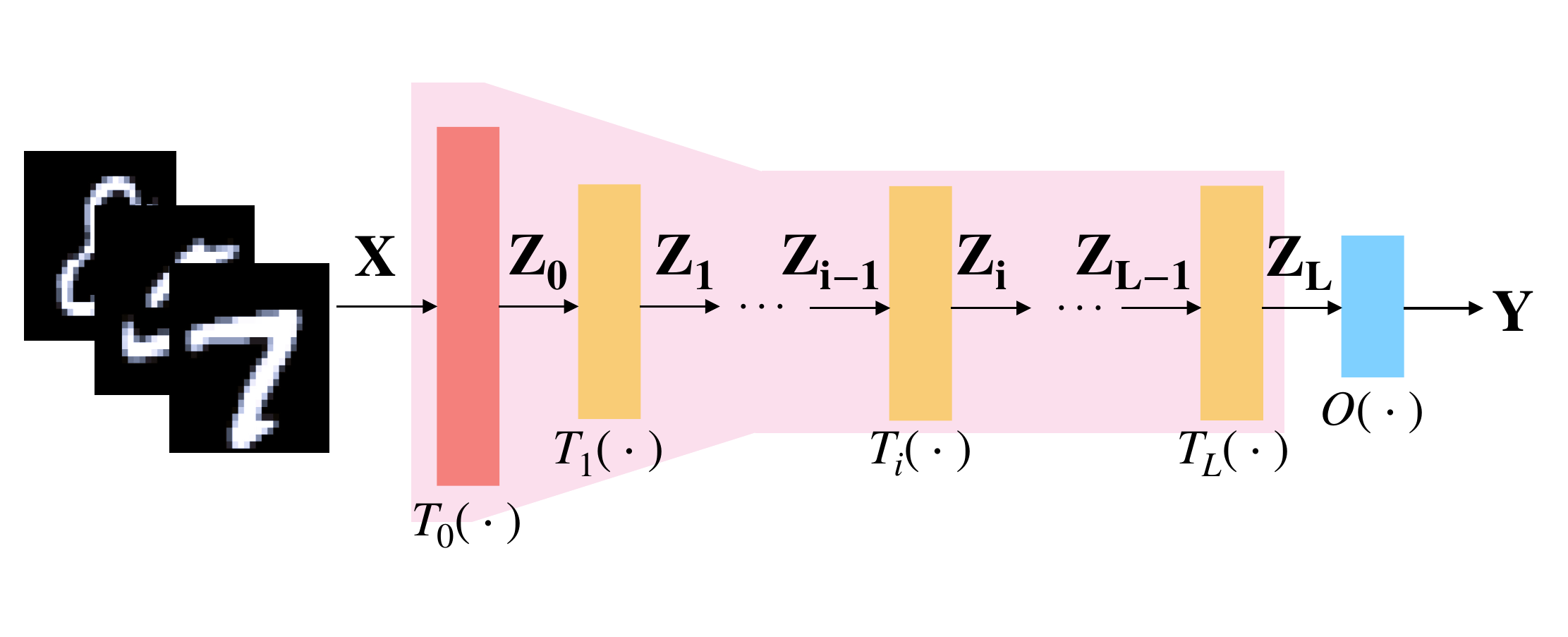}
    \caption{\PRETRAINED\ Network}
    \label{img:pt:1}
  \end{subfigure}
  \hspace{1mm}
  \begin{subfigure}[b]{\columnwidth} 
    \centering
    \includegraphics[width=\columnwidth]{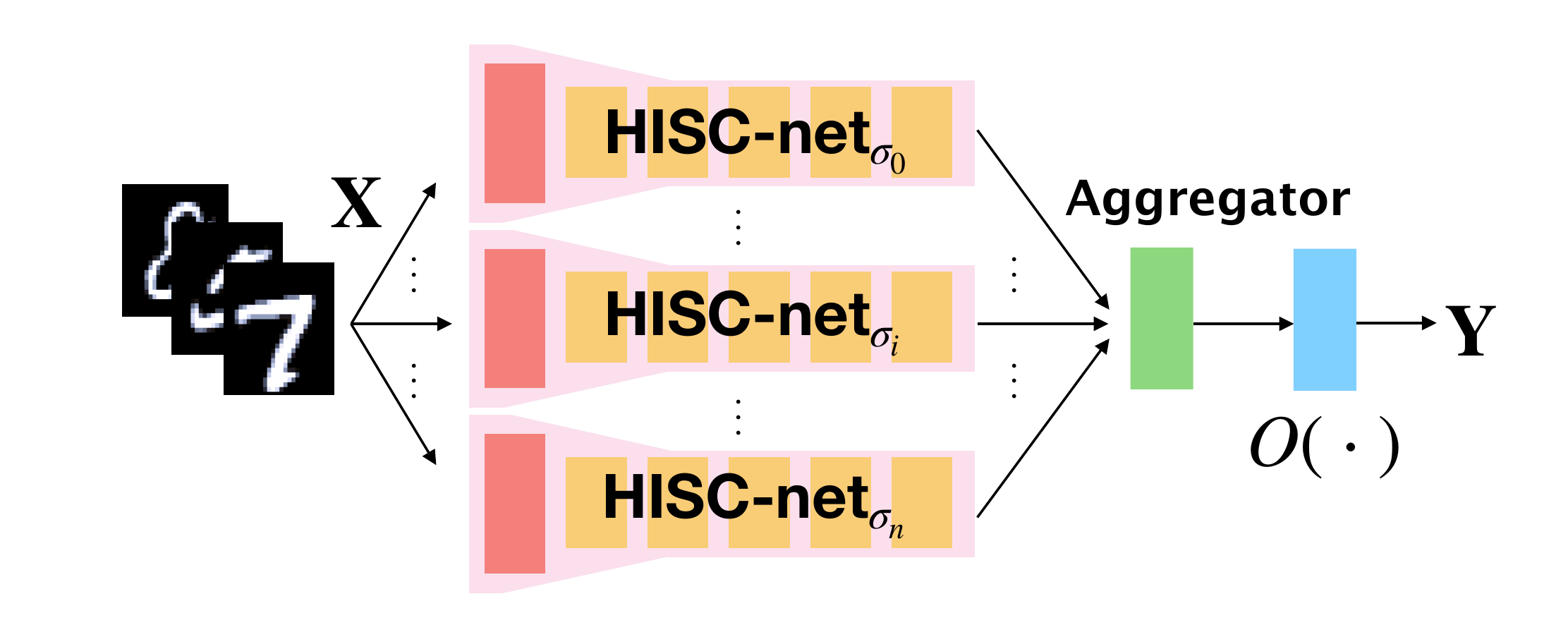}
    \caption{\COMBINEDNET}
    \label{img:pt:2}
  \end{subfigure}
  \caption{The \PRETRAINED\ network (\Fig~\ref{img:pt:1}) is a standard feedforward network trained using the HSIC-bottleneck objective, resulting in hidden representations at the last layer that can easily be used for classification.
  \Fig~\ref{img:pt:2} shows a \COMBINEDNET, where each branch $\text{HSIC-net}_{\sigma_j}$ is trained with a specific $\sigma$. The aggregator averages the hidden representations to form an output representation.}
  \label{img:pt}
\end{figure}

\subsection{HSIC-Bottleneck} \label{ss:pt}
Suppose we have a network composed of $L$ hidden layers $T_{i}(\cdot): \mathbb{R}^{d_{i-1}} \rightarrow \mathbb{R}^{d_{i}}$, resulting in hidden representations\footnote{This can be extended to the activations produced from convolutional layers, as each activation is flattened and stacked in array $\ZZ_i$.} 
$\ZZ_i \in \mathbb{R}^{m \times d_{i}}$, where $i \in \{1,...,L\}$, and $m$ denotes batch size. Implementing the Information Bottleneck principle, we replace the original mutual information terms with nHSIC \eqref{eq:nhsic} as the learning objective:
\begin{equation}
 \label{eq:hsic01}
  \ZZ_i^{\ast} = \argmin_{\ZZ_i} \ \nHSIC(\ZZ_i, \mathbf{X}) - \beta \, \nHSIC(\ZZ_i, \YY), 
\end{equation}
where $\XX \in \mathbb{R}^{m \times d_x}$ is the input, $\YY \in \mathbb{R}^{m \times d_y}$ is the label, $i \in \{0,...,L\}$ and $L$ is the number of hidden layers, $d_x$ and $d_y$ are the dimensionalities of the input and output variables respectively. 
Since we concentrate on classification in our experiments, $d_y$ is the number of classes (we expect the HSIC bottleneck approach can be applied to other tasks such as regression).
The $\beta$ controls the balance of IB objectives. 
Following nHSIC of \eqref{eq:nhsic}, the nHSIC of each term is:

\begin{equation}
  \label{eq:hsic02}
  \begin{split}
    \nHSIC(\ZZ_i, \XX) = \tr(\widetilde{\KK}_{Z_i}\widetilde{\KK}_X)
    \\
    \nHSIC(\ZZ_i, \YY) = \tr(\widetilde{\KK}_{Z_i}\widetilde{\KK}_Y)
  \end{split}
\end{equation}

The formulation \eqref{eq:hsic01}, \eqref{eq:hsic02} suggests that the optimal hidden representation $\ZZ_{i}$ finds a balance between independence from unnecessary details of the input and dependence with the output. Ideally, the information needed to predict the label is retained when \eqref{eq:hsic01} converges, while unnecessary information that would permit overfitting is removed. 

We optimize equation \eqref{eq:hsic01} independently at each layer using block coordinate descent without gradient propagation.

\begin{algorithm}
\DontPrintSemicolon
\KwData{
    $\XX_j \in \mathbb{R}^{m \times d_x}$: data batch $j$,
    $\YY_j \in \mathbb{R}^{m \times d_y}$: label batch $j$,
    layer $T_i$ parameterized by $\{\bm{\theta}_i|\WW_i, \bb_i\}$, 
    $i \in \{1,...,L\}$: layer iterator, 
    $j \in \{1,..., \floor{n/m}\}$: batch iterator, 
    $m$: batch size, 
    $n$: number of input data,
    $\alpha$: learning rate.
}
\KwResult{Saved HSIC-trained network $T_i(\cdot): \mathbb{R}^{d_{i-1}} \rightarrow  \mathbb{R}^{d_i}$, $i \in \{1,...,L\}$}
 \For{$j \in \{1,...,n/m\}$}
 {
   \For{$i \in \{1,...,L\}$}
   {
    $\ZZ_i = \TT_{i-1}(\ZZ_{i-1})$ $\quad\quad$ \tcp{$\ZZ_0 = \XX_j$}
    $\bgg_i = \nabla_{\bm{\theta}_i} (\nHSIC(\ZZ_i, \XX_j) - \beta \nHSIC(\ZZ_i, \YY_j))$\;
    $\bm{\theta}_i \leftarrow \bm{\theta}_i - \alpha \bgg_i$ \;  
   }
 }
 \caption{Unformatted training}
 \label{algo:hsicbt}
\end{algorithm}
\subsection{\UNFORMATTEDC} \label{ss:pn}

As shown in the Experiments section,
HSIC-bottleneck training tends to produce one-hot outputs in some experiments. This inspired us to use the HSIC-bottleneck objective to directly solve the classification problem. This is done by setting the dimensionality of the last layer $\ZZ_L$ to match the number of classes (e.g., $d_L = d_y$). Since the resulting activation is typically permuted with respect to the labels (e.g., images of the digit zero might activate the forth output layer entry), we simply find a fixed permutation by picking the output with the highest activation across the inputs of a particular class as the output for that class. Please refer to Algorithm~\ref{algo:hsicbt} for more detail.

\subsection{\POSTTRAINEDC\ network} 

We append a single output layer $O(\cdot): \mathbb{R}^{d_L} \rightarrow \mathbb{R}^{d_y}$  equipped with a softmax function for classification (\Fig~\ref{img:pt}), taking the optimized last-layer hidden representation 
from a \UNFORMATTEDED\ network as its input. The new layer is trained using minibatch SGD with the loss
\[
  L_{\text{format}} = \text{CE}\big(\YY, O(\ZZ_L)\big)   
\] 
where CE denotes the cross entropy loss.
\subsection{\COMBINEDNETC} \label{ss:sn}	

In principle, HSIC is a powerful measure of statistical independence, however in practice the results do depend somewhat on the chosen $\sigma$ parameter even when using the normalized form \eqref{eq:nhsic}. To cope with this, we combine \UNFORMATTEDED\ networks with different $\sigma$, and then aggregate the resulting hidden representations.
This \COMBINEDNET\ architecture is illustrated in \Fig~\ref{img:pt:2}, and has the objective
\[
    L_{\text{Comb}}(\XX, \YY) =  \text{CE}\bigg({\YY,O\big(\frac1n \sum_i^n \text{HSIC-net}_{\sigma_i}(\XX)\big)}\bigg)
\]
where the output classifier layer $O(\cdot)$ takes the average of  representations from the \UNFORMATTEDED\ networks, $\text{HSIC-net}_{\sigma_i}$, trained with $\sigma_i$, $i \in \{1,...,n\}$. Then it optimizes the layer $O(\cdot)$ with cross entropy while keeping the trained $\text{HSIC-net}_{\sigma_i}$ fixed. The performance of the \COMBINEDNET\ is presented below in Section~\ref{ss:cap}.

\subsection{Performance} 		
Performance comparisons with backpropagation are difficult since the HSIC bottleneck performance depends heavily on the minibatch size. SGD backpropagation schemes are linear in the number of data points.  HSIC is $O(m^2)$ where $m$ is the minibatch size in our case.\footnote{However our github reference code uses an $O(m^3)$ implementation for simplicity.}
Typically the minibatch size is a constant that is chosen based on validation performance and/or available GPU memory rather than scaling with the data, so strictly speaking the  HSIC-bottleneck approach is also linear in the number of data points.
However, the learning convergence is a quantity of ultimate interest. This quantity is not known for either backpropagation or HSIC bottleneck, and it may be different considering the fundamentally distinct character of these two approaches. HSIC bottleneck is more amenable to layer parallel computation, since the need for backpropagation is removed.

\section{Experiments} \label{s:experiments}
In this section, we report several experiments that explore and validate the \PRETRAINED\ network concept. First, to motivate our work, we plot the HSIC-bottleneck values and activation distributions of a simple model during backpropagation training. 
We then show how  \UNFORMATTED\ can produce one-hot results that are directly ready for classification with a shallow or deep network.
Next, we compare backpropagation with \POSTTRAINING\ on networks with different numbers of layers.
The value of \PRETRAINING\ for \POSTTRAINING\ and the effect of the hyperparameter $\sigma$ are considered in the next experiments.  Lastly, we briefly consider the application of \PRETRAINING\ to other network architectures such as ResNet. 

For the experiments, we used standard feedforward networks with batch-normalization \cite{IoffeS15} on the MNIST/Fashion MNIST/CIFAR10 datasets. All experiments including standard backpropagation, \UNFORMATTEDED, and \POSTTRAINED, use a simple SGD optimizer. The coefficient $\beta$ and the kernel scale factor $\sigma$ of the HSIC-bottleneck were set to 500 and 5 respectively,\footnote{For complete details of the experiments refer to our github repository.} which empirically balances  compression and the relevant information available for the classification task. 

\begin{figure}[ht]
  \centering
  \begin{subfigure}[b]{0.32\linewidth} 
    \includegraphics[width=\linewidth]{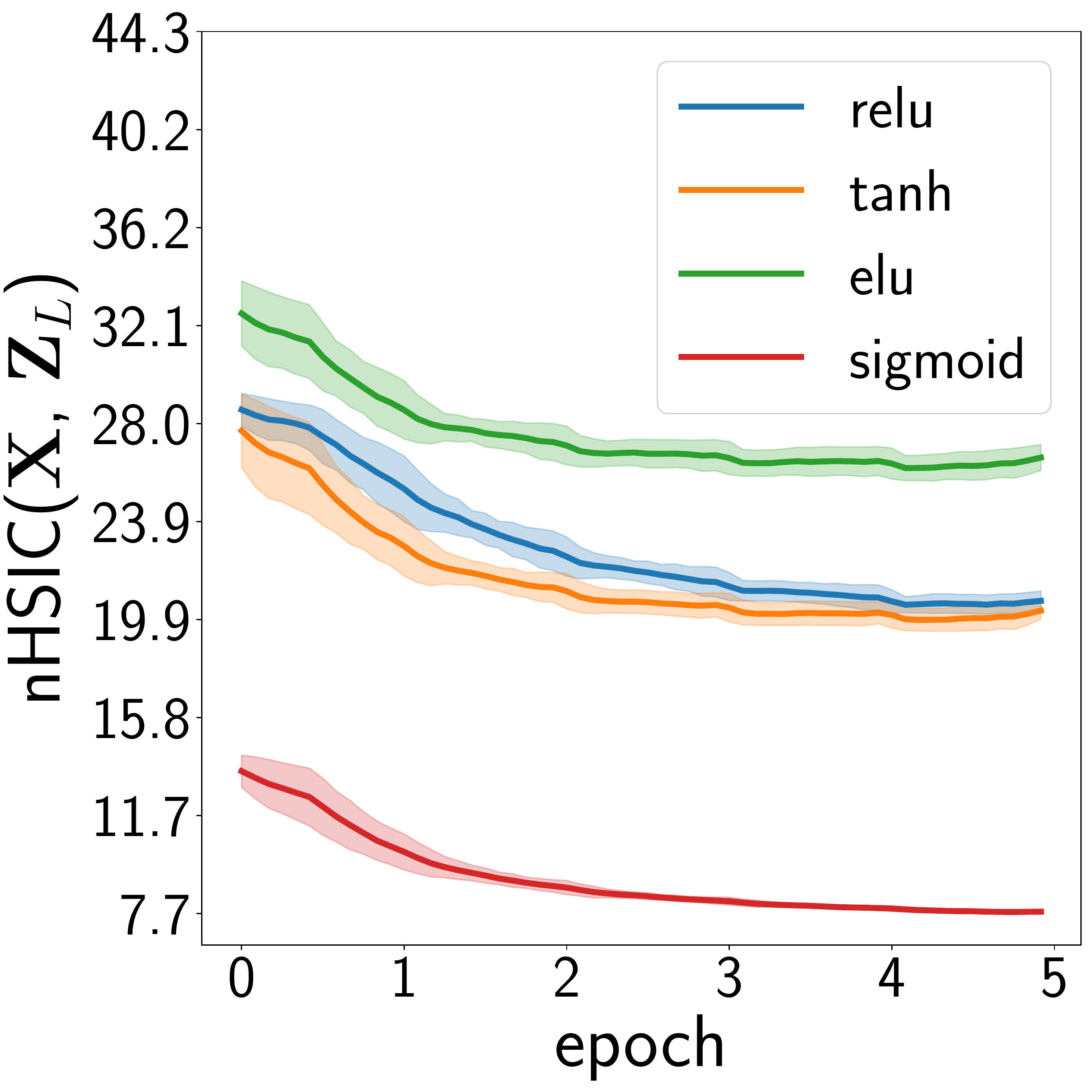}
    \caption{}
    \label{img:hbdna:1}
  \end{subfigure}
  \begin{subfigure}[b]{0.32\linewidth} 
    \includegraphics[width=\linewidth]{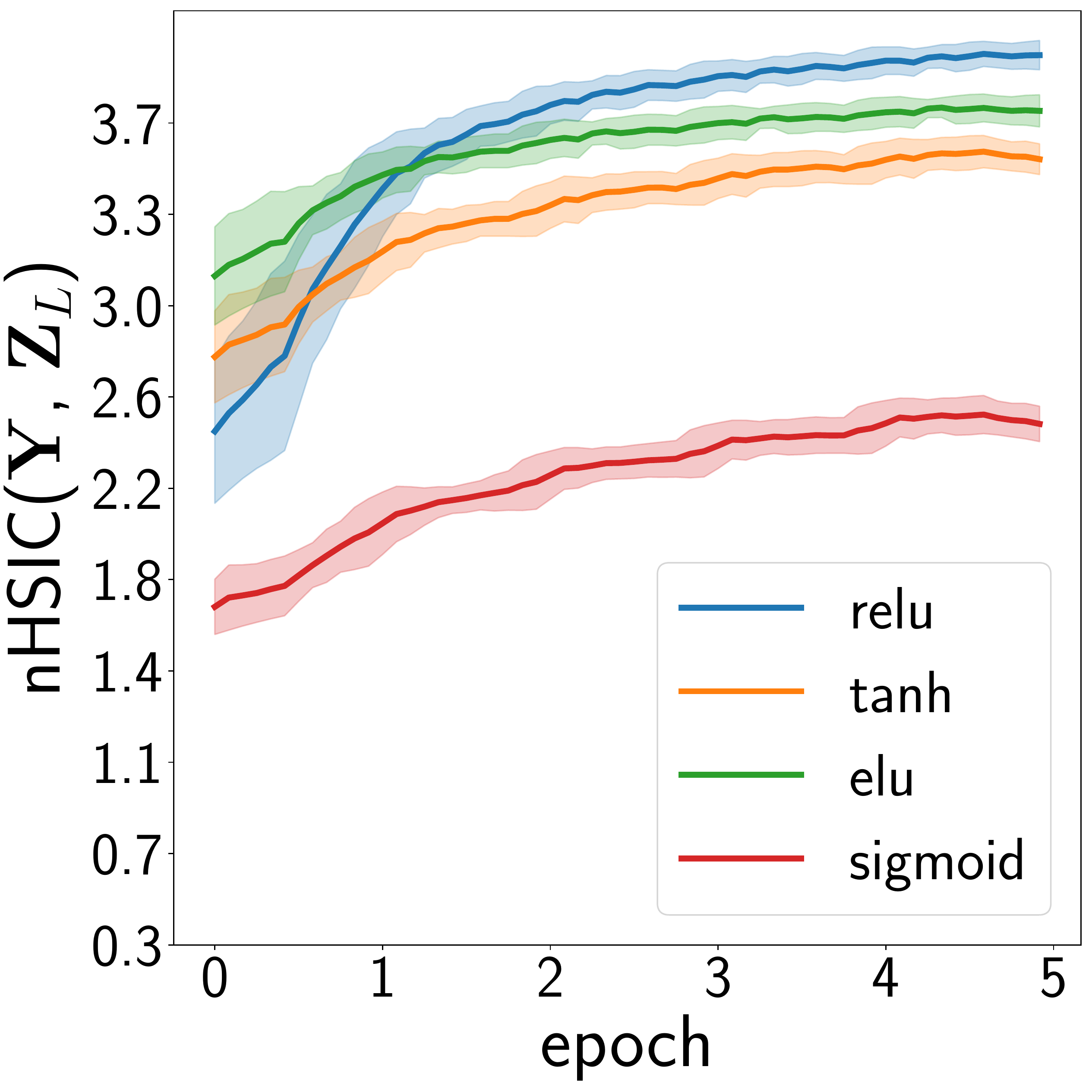}
    \caption{}
    \label{img:hbdna:2}
  \end{subfigure}
  \begin{subfigure}[b]{0.32\linewidth} 
    \includegraphics[width=\linewidth]{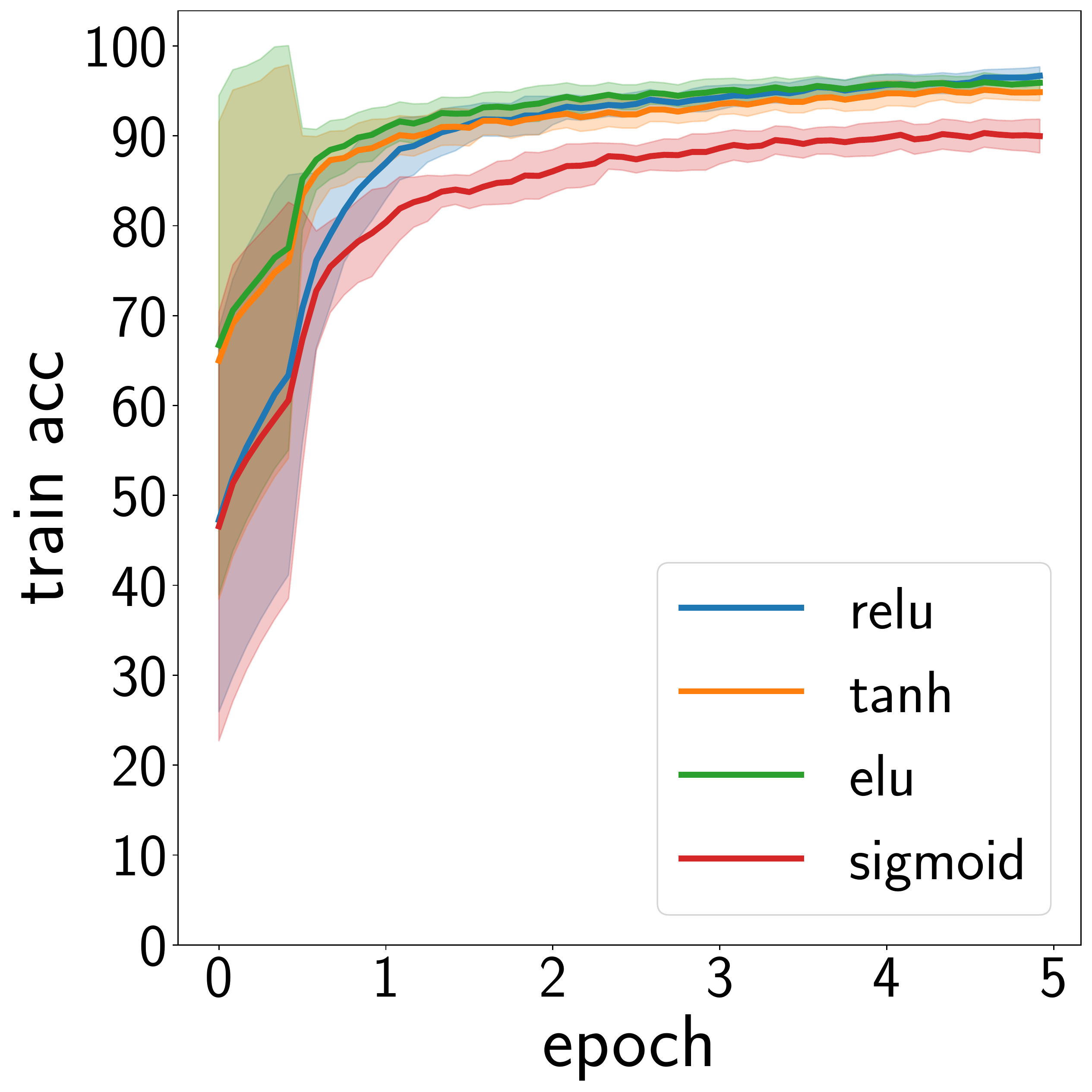}
    \caption{}
    \label{img:hbdna:3}
    
  \end{subfigure}
    \begin{subfigure}[b]{0.32\linewidth} 
    \includegraphics[width=\linewidth]{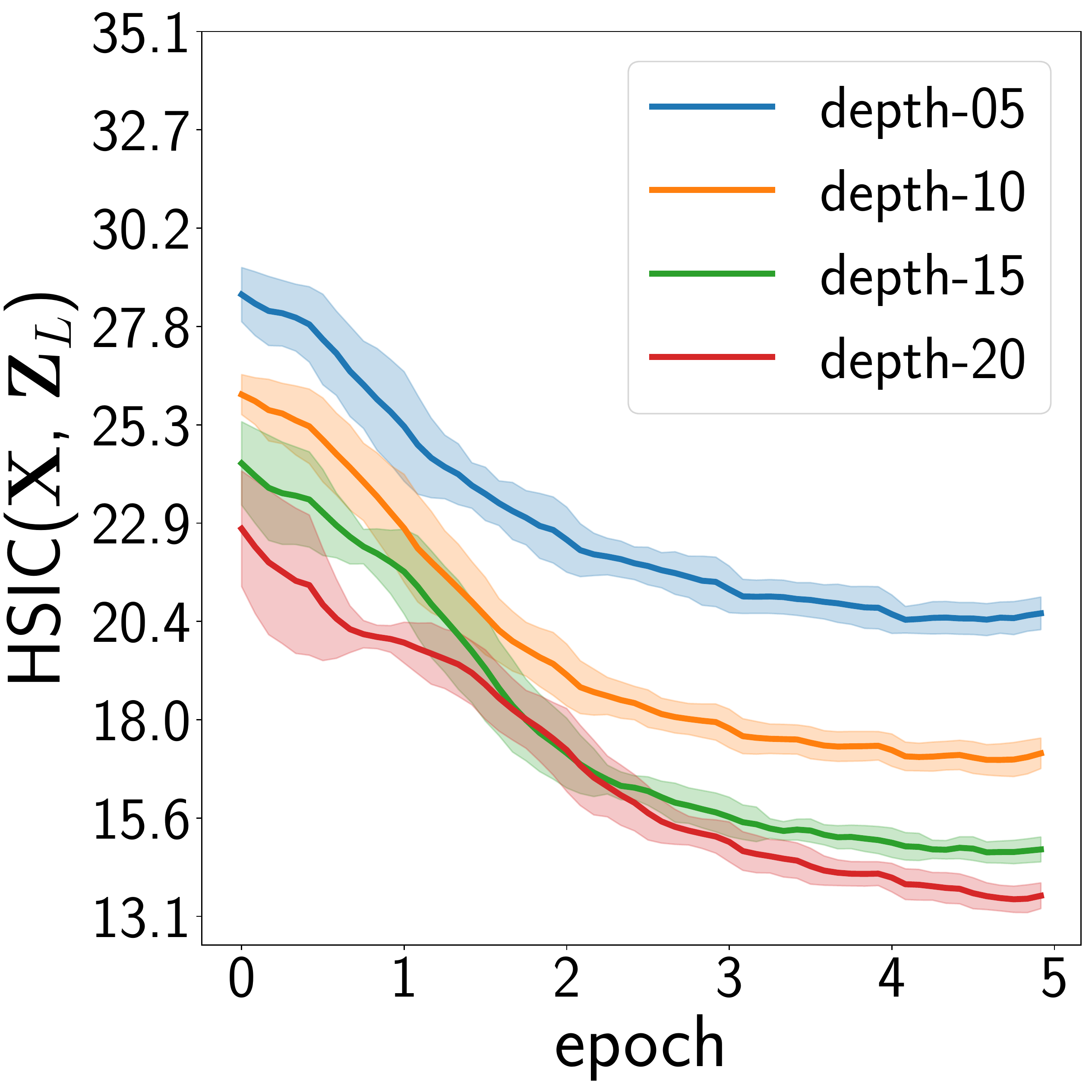}
    \caption{}
    \label{img:hbdna:4}
  \end{subfigure}
  \begin{subfigure}[b]{0.32\linewidth} 
    \includegraphics[width=\linewidth]{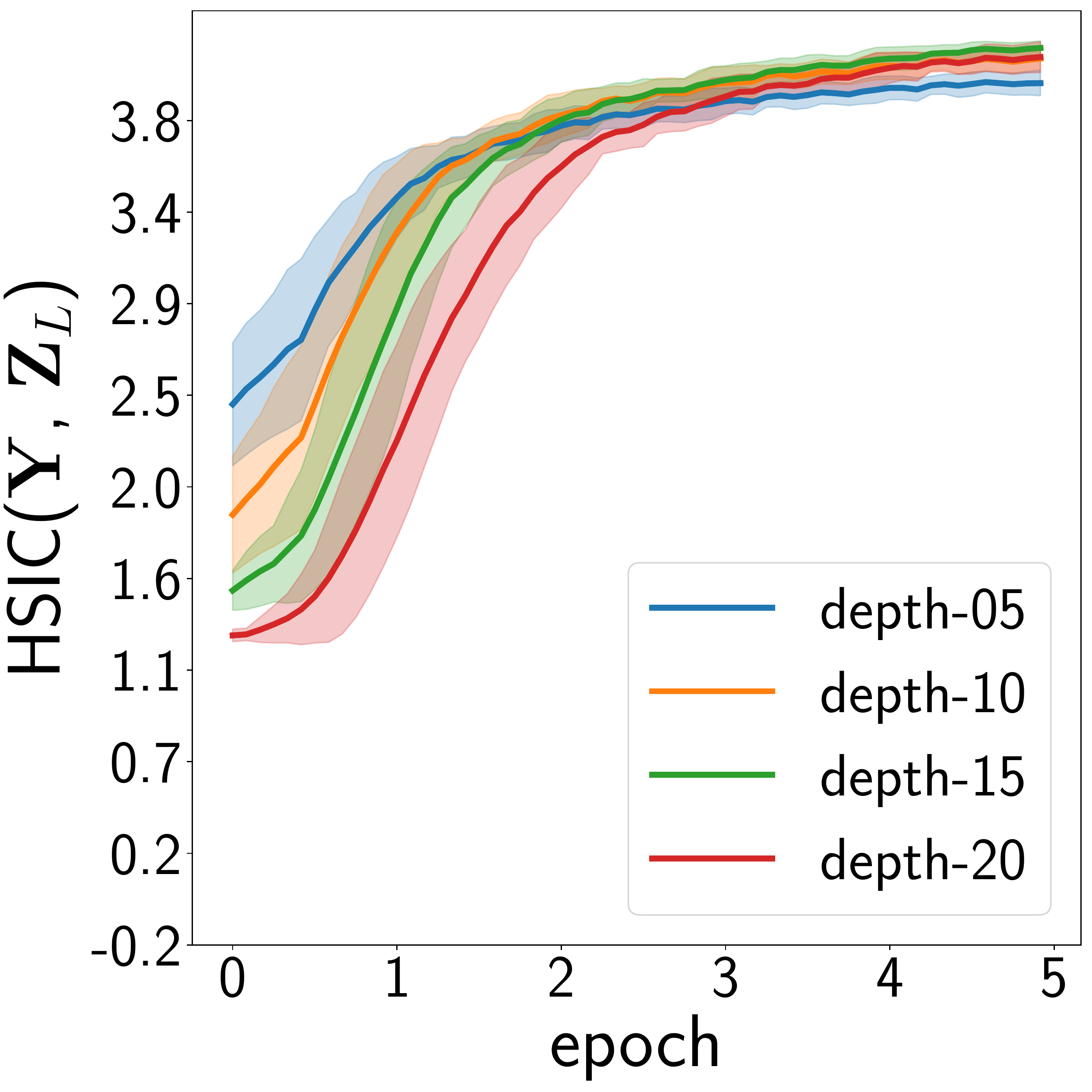}
    \caption{}
    \label{img:hbdna:5}
  \end{subfigure}
  \begin{subfigure}[b]{0.32\linewidth} 
    \includegraphics[width=\linewidth]{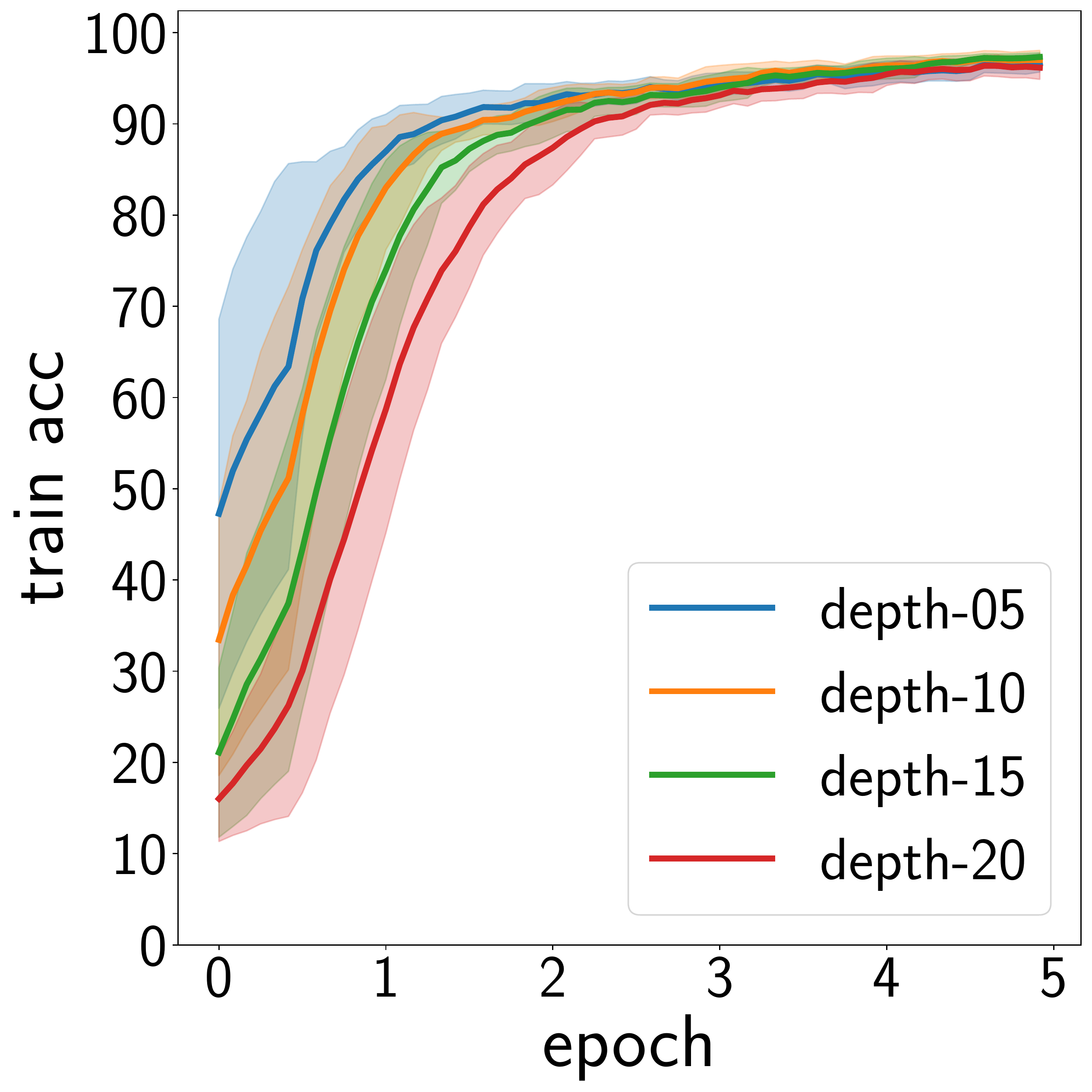}
    \caption{}
    \label{img:hbdna:6}
  \end{subfigure}
   \caption{Visualization of the HSIC-bottleneck quantities $\nHSIC(\XX,\ZZ_L)$, $\nHSIC(\YY,\ZZ_L)$ and training accuracy as a function of epoch number, monitored during conventional \emph{backpropagation} training across various network activation functions \Figs~(\ref{img:hbdna:1})-(\ref{img:hbdna:3}) and depths \Figs~(\ref{img:hbdna:4})-(\ref{img:hbdna:6}). \Figs~(\ref{img:hbdna:4})-(\ref{img:hbdna:6}) use ReLU with different network depths. \Figs~(\ref{img:hbdna:1})--(\ref{img:hbdna:3}) vary only the activation type. The shaded area in the plots represents the standard deviation of the unsmoothed training performance.}
  \label{img:hbdna}
\end{figure}

Before considering the use of the HSIC-bottleneck as a training objective, we first  validate its relevance in the context of conventional deep network training (\Fig~\ref{img:hbdna}).
Monitoring the nHSIC between hidden activations and the input and output of a simple network trained using backpropagation shows that $\nHSIC(\YY,\ZZ_L)$ rapidly increased during early training as representations are formed, while $\nHSIC(\XX,\ZZ_L)$ rapidly drops.
The value of $\nHSIC(\YY,\ZZ_L)$ varies with the network depth (\Fig~\ref{img:hbdna:5}) and depends on the choice of activation (\Fig~\ref{img:hbdna:2}). Furthermore, it clearly parallels the increase in training accuracy (\Fig~\ref{img:hbdna:3}, \Fig~\ref{img:hbdna:6}). In summary, \Fig~\ref{img:hbdna} shows that a range of different networks follow the information bottleneck principal.

\subsection{\UNFORMATTEDC} \label{ss:hbdna}

\begin{figure}[ht]
  \centering
  \hspace{1mm}
  \includegraphics[width=\linewidth]{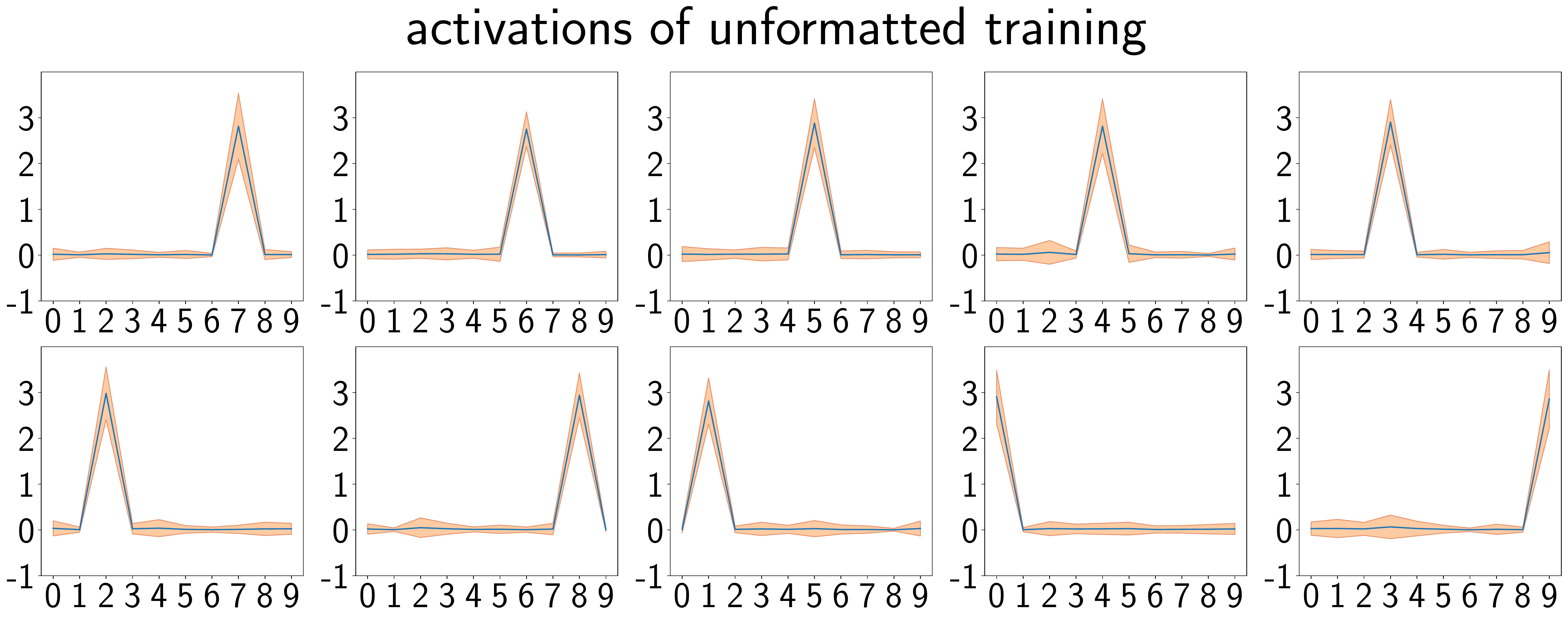}
   \caption{The MNIST output category distribution for a shallow fully-connected network. Each of the subfigures is the output for a specific category, from the `zero' digits (top-left) to the `nine' digits (bottom right). The zero inputs produced a one-hot activation in position seven, and the nines produces a one-hot activation in position nine, respectively. The particular permutation depends on the random weight initialization.}
  \label{img:presolveonehot}
\end{figure}

Next, we try using the HSIC-bottleneck as the sole training objective. We use a fully connected network architecture 784-256-256-256-256-256-10 with ReLU activation functions.
Remarkably, in experiments on CIFAR10, FashionMNIST, and MNIST, the HSIC-trained network often results in non-overlapping one-hot output activations for many (but not all) random weight initializations, as seen in \Fig~\ref{img:presolveonehot}. This allows classification to be performed by simply using a fixed permutation, i.e.,~using the highest activation distribution value to select the class. For example in \Fig~\ref{img:presolveonehot}, the class activities of MNIST dataset from digit zero to nine have highest density at entries: 7, 6, 5, 4, 3, 2, 8, 1, 0, 9. We also tested a toy model to show that a network trained with \UNFORMATTED\ can separate classes in a $\mathbb{R}^1$ latent space. More detail on this experiment is shown in the Appendix.

\begin{figure}[ht]
  \centering
  \begin{subfigure}[b]{0.32\linewidth} 
    \includegraphics[width=\linewidth]{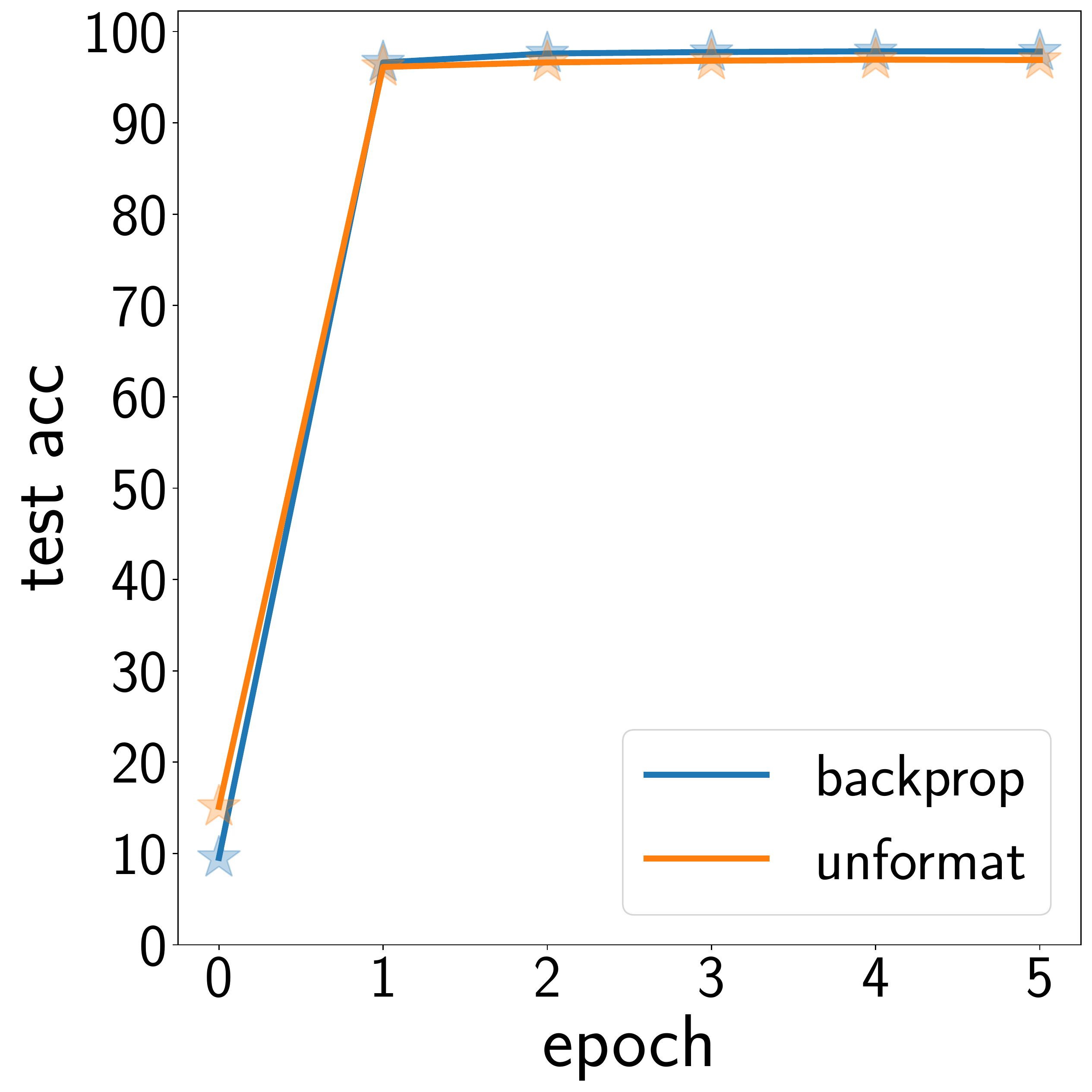}
    \caption{MNIST}
  \end{subfigure}
  \begin{subfigure}[b]{0.32\linewidth} 
    \includegraphics[width=\linewidth]{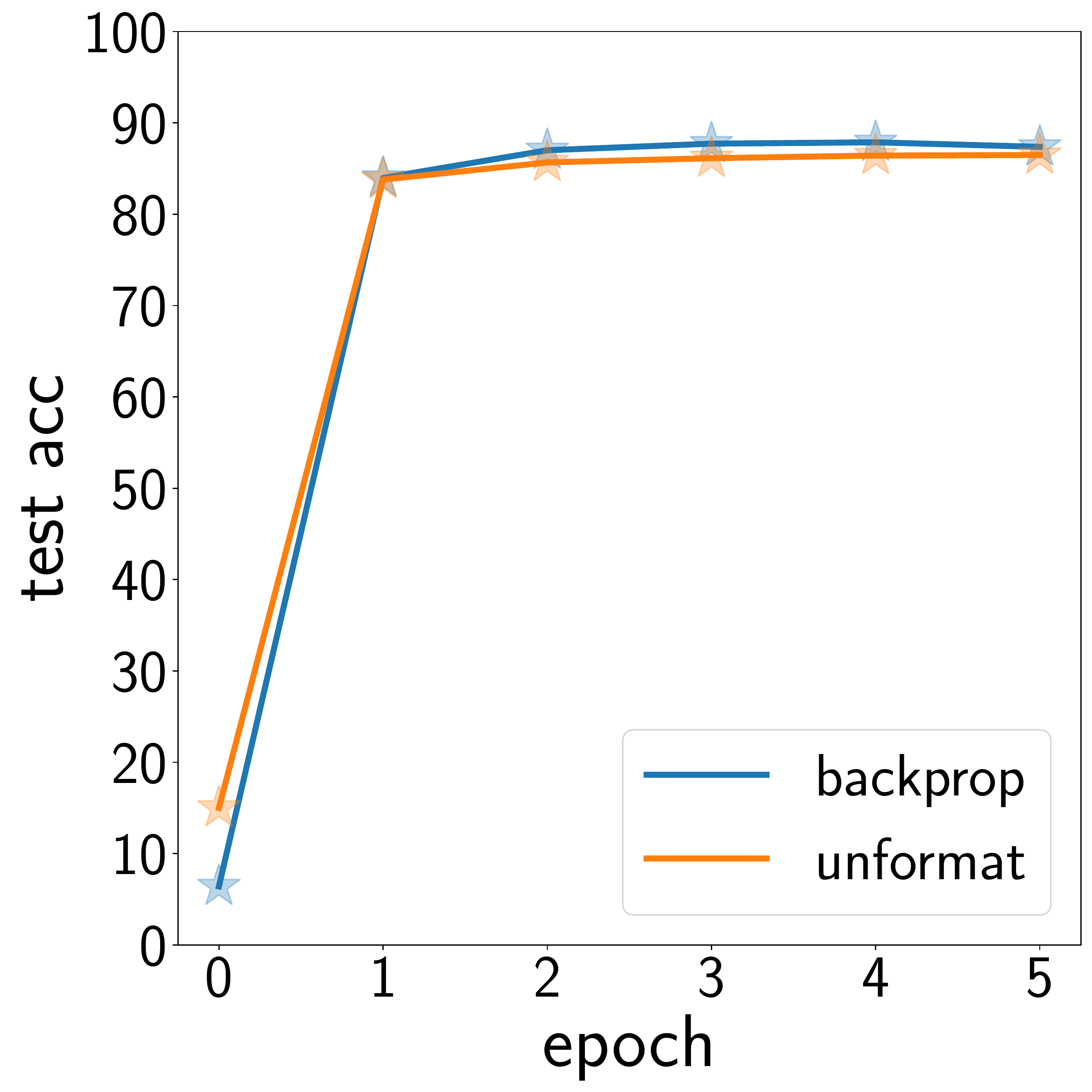}
    \caption{FashionMNIST}
  \end{subfigure}
  \begin{subfigure}[b]{0.32\linewidth} 
    \includegraphics[width=\linewidth]{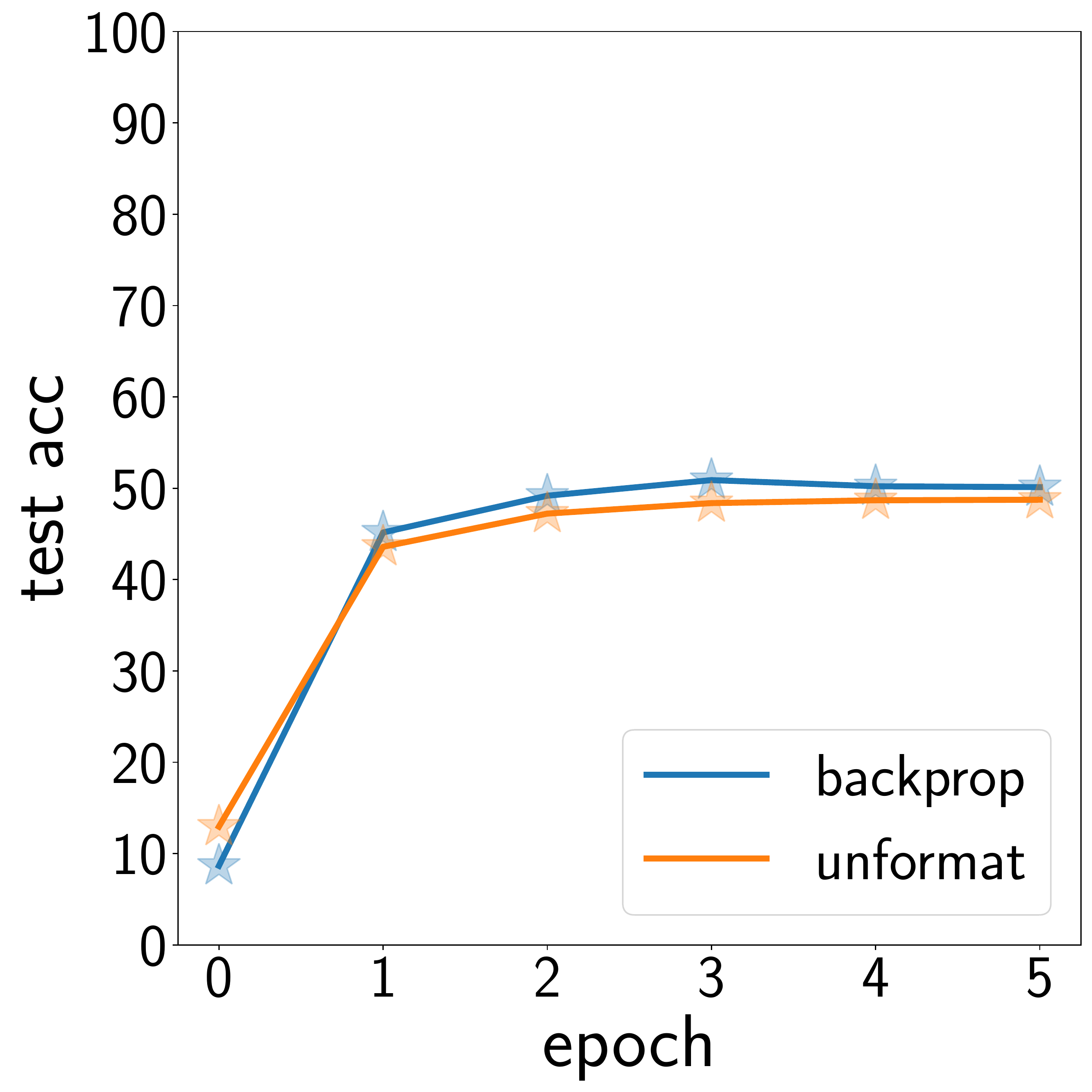}
    \caption{CIFAR10}
  \end{subfigure}
    \begin{subfigure}[b]{0.32\linewidth} 
    \includegraphics[width=\linewidth]{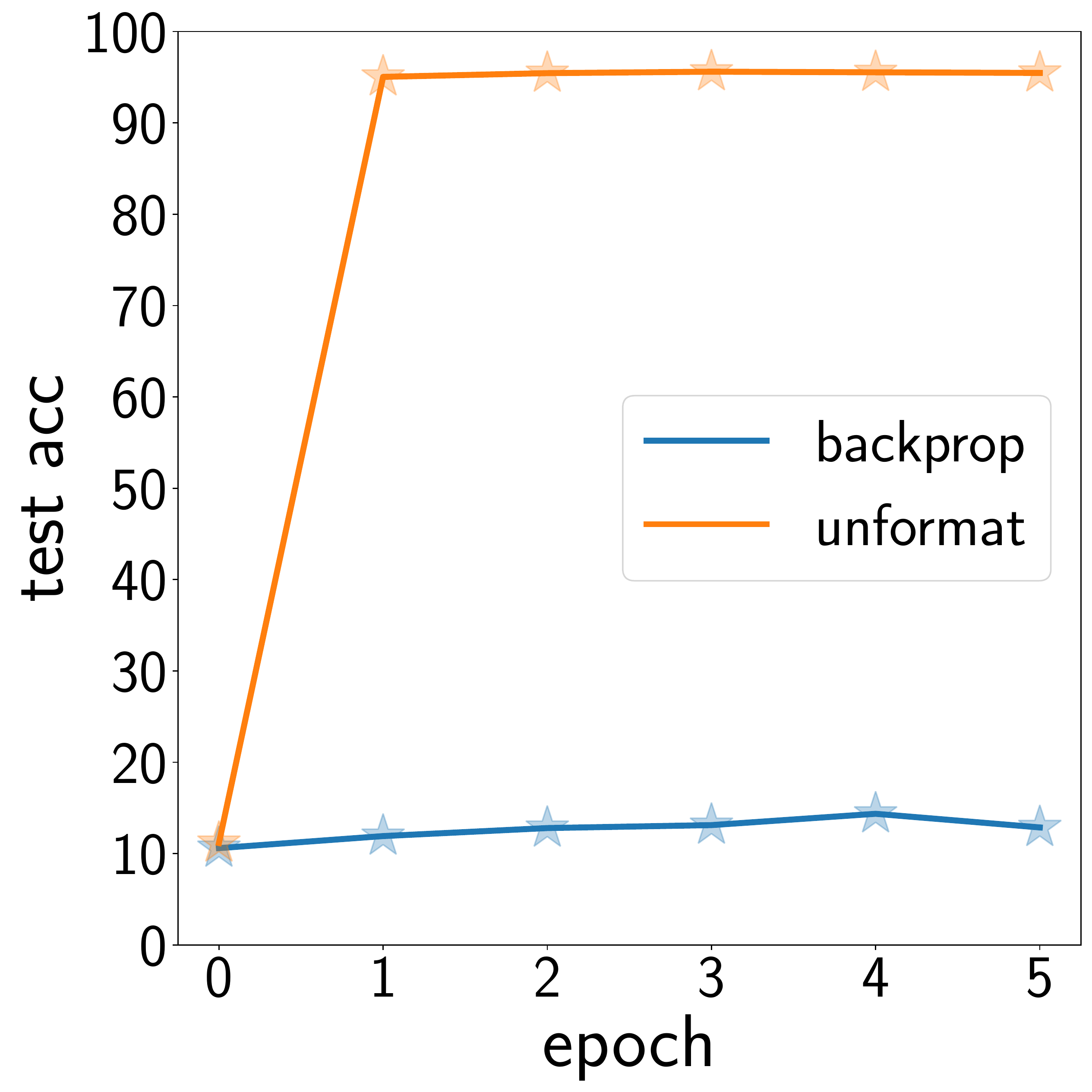}
    \caption{MNIST}
  \end{subfigure}
  \begin{subfigure}[b]{0.32\linewidth} 
    \includegraphics[width=\linewidth]{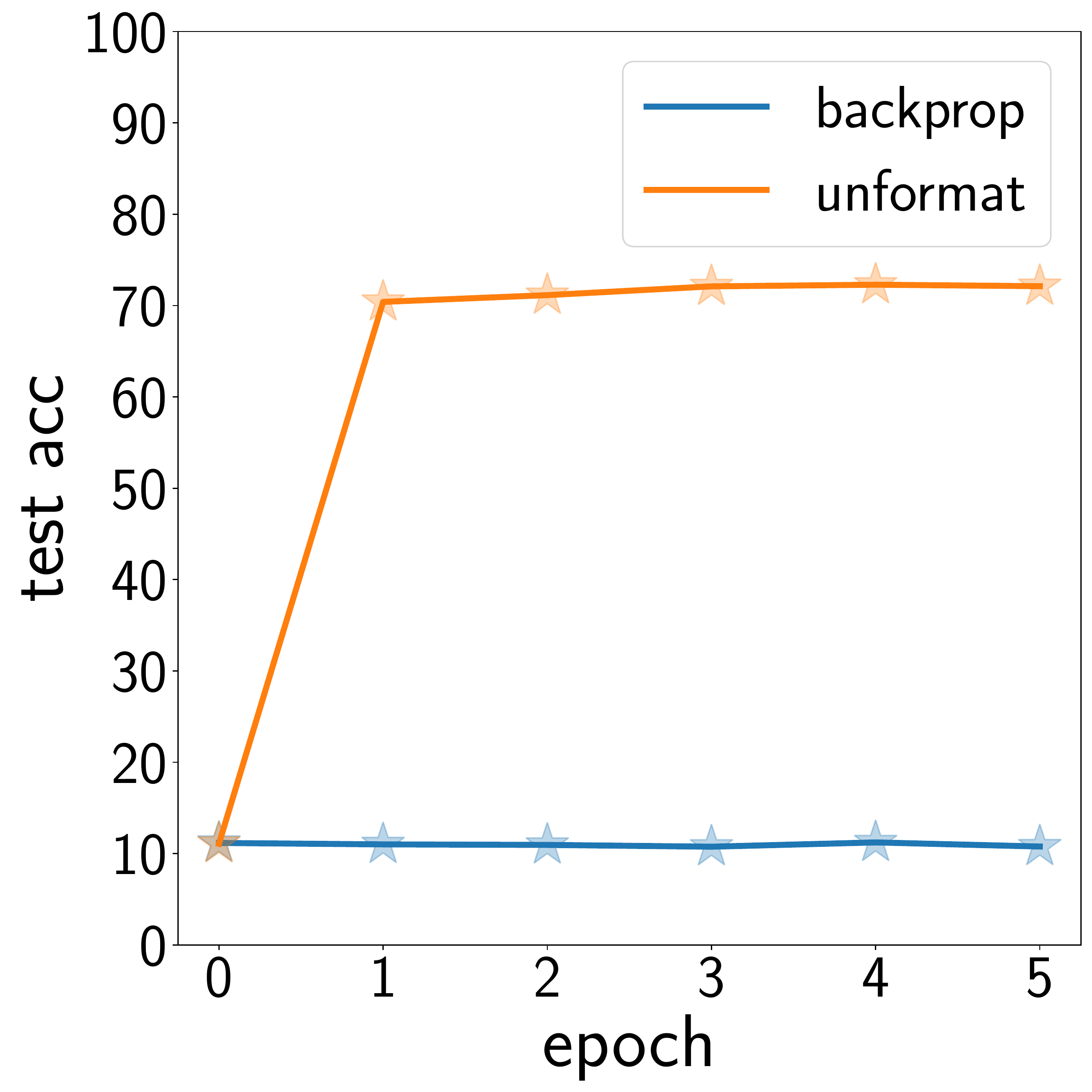}
    \caption{FashionMNIST}
  \end{subfigure}
  \begin{subfigure}[b]{0.32\linewidth} 
    \includegraphics[width=\linewidth]{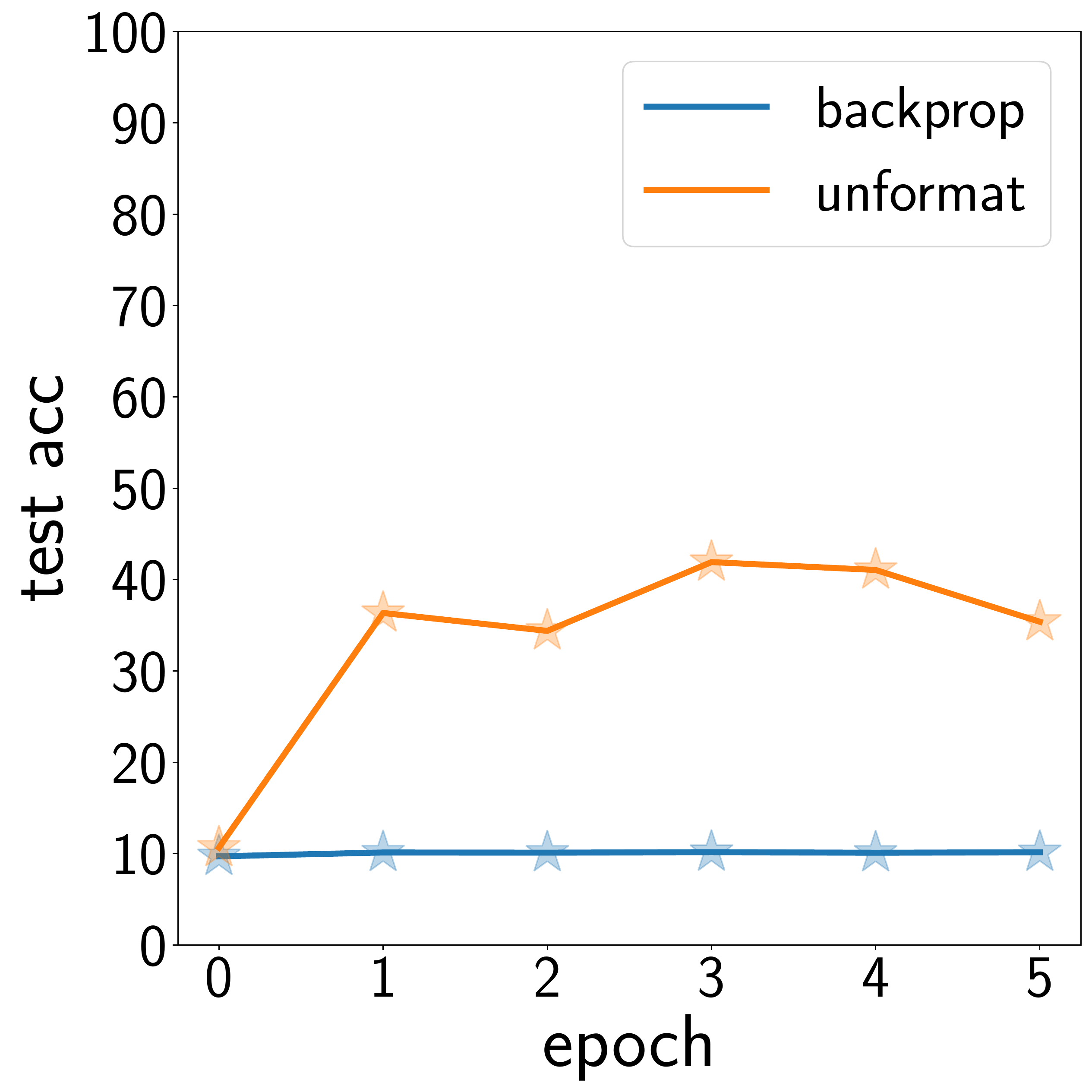}
    \caption{CIFAR10}
  \end{subfigure}
   \caption{The test accuracy of HSIC-bottleneck training on standard classification problems as a function of epoch number. The experiments use the same architecture where the top- and bottom-row use 5 and 50 hidden layers, respectively.}
  \label{img:presolve}
\end{figure}

Our approach can produce results competitive with standard training. \Fig~\ref{img:presolve} illustrates the result for both backpropagation and \UNFORMATTED\ for otherwise identical networks. 
We see that \UNFORMATTED\ provides comparable results to backpropagation when the network is shallow (top-row); however, the performance of backpropagation is poor for deep networks (bottom-row). Note that the CIFAR10 results use a fully connected rather than convolutional network.

The obtained results support the idea that \UNFORMATTED\  encodes the input variables in a form from which the desired output can be easily discovered, either by simple permutation or by \POSTTRAINEDING.

\subsection{Format Training Results}\label{ss:ssi}

\begin{figure}[ht]
  \centering
  \begin{subfigure}[b]{0.45\linewidth} \label{fig_1}
    \includegraphics[width=\linewidth]{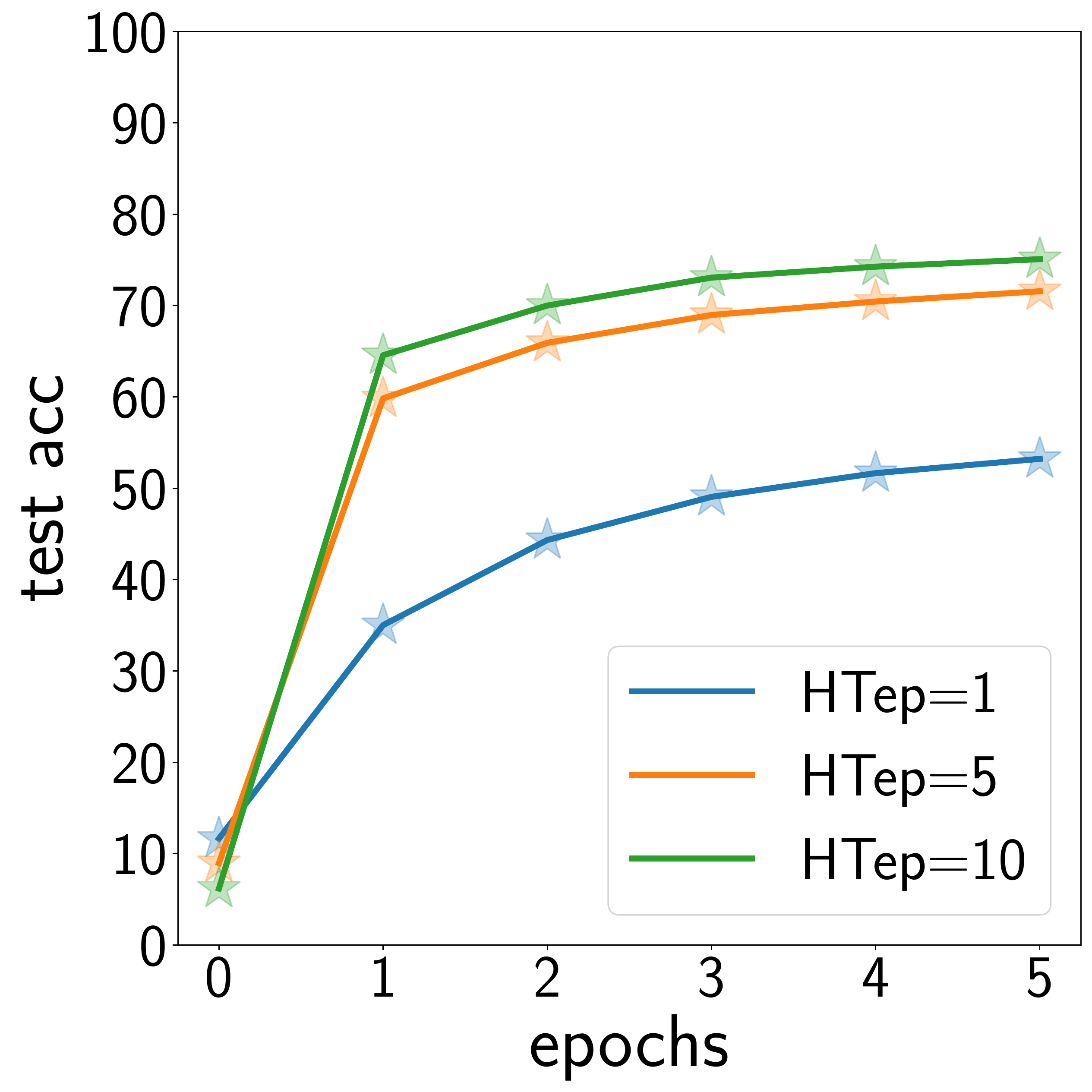}
    \caption{training accuracy}
  \end{subfigure}
  \hspace{1mm}
  \begin{subfigure}[b]{0.45\linewidth} \label{fig_2}
    \includegraphics[width=\linewidth]{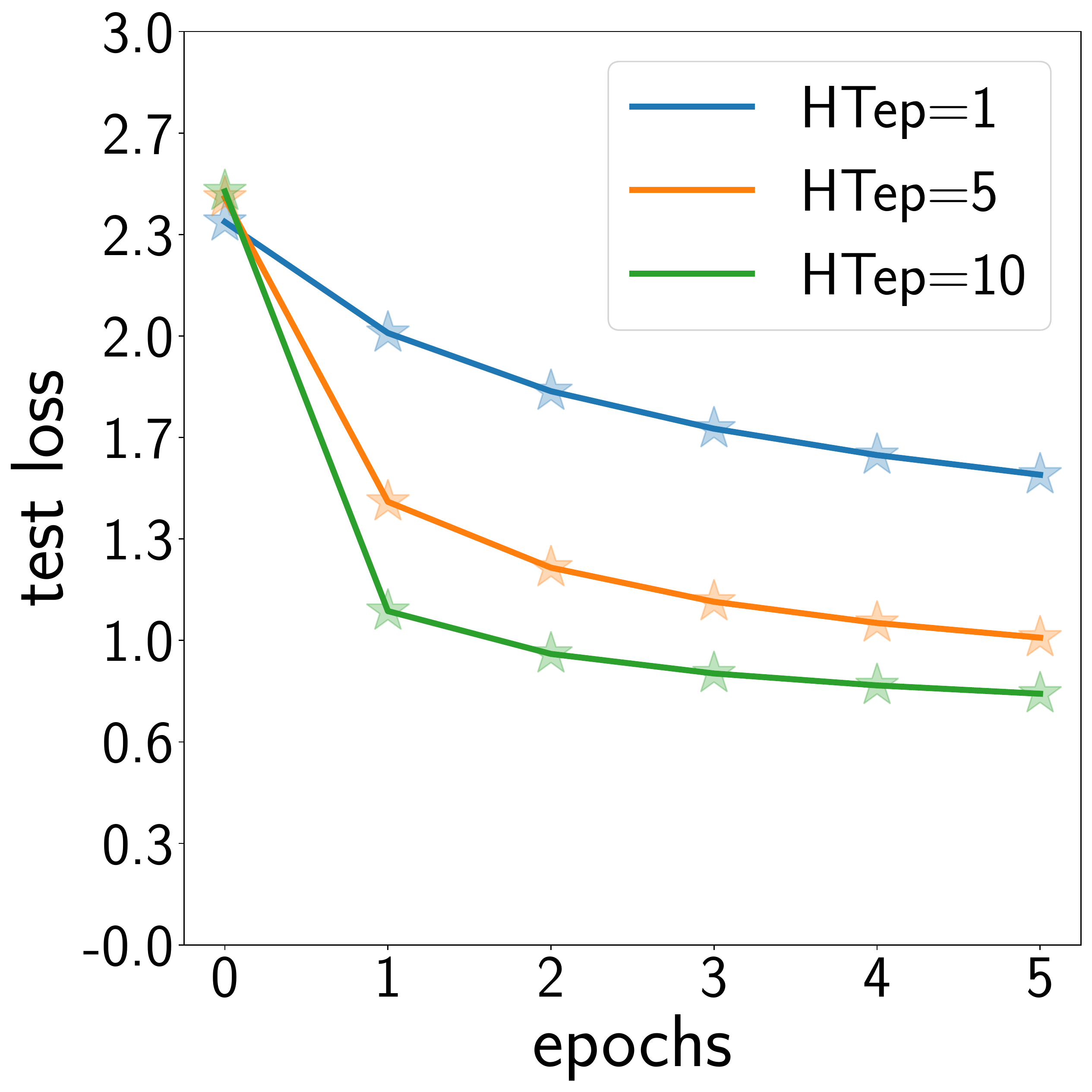}
    \caption{training loss}
  \end{subfigure}
  \caption{\POSTTRAINEDC\ training accuracy as a function of epoch number for a  \UNFORMATTEDED network trained for 1, 5, and 10 epochs. ``HTep'' denotes the number of epochs that the  \UNFORMATTEDED network used. \UNFORMATTEDC\ to convergence provides better \POSTTRAINING\ performance in this experiment. 
  }
  \label{img:ssi}
\end{figure}

An interesting question regarding deep neural networks is how effectively these stacked layers learn the information from the input and the label. To explore this, we fixed all the hyper-parameters of an \UNFORMATTEDED\ network except the training time (number of epochs). We expect that training the  \UNFORMATTEDED\ network for more epochs will result in a hidden representation that better represents the information needed to predict the label, resulting in higher accuracy in the \POSTTRAINEDING\ stage. \Fig~\ref{img:ssi} shows the result of this experiments, specifically, the accuracy and loss of \POSTTRAINEDING\  on a five-layer  \UNFORMATTEDED\ network trained with 1, 5, and 10 epochs. From \Fig~\ref{img:ssi} it is evident that the  \UNFORMATTEDED\ network can boost accuracy at the beginning of SGD \POSTTRAINING. Additionally, as the  \UNFORMATTEDED\ network trains longer, the \POSTTRAINING\ yields higher accuracy. 

\subsection{Network Capacity and Scale} \label{ss:cap}

\begin{figure}[ht]
  \centering
  \begin{subfigure}[b]{0.31\linewidth}
    \includegraphics[width=\linewidth]{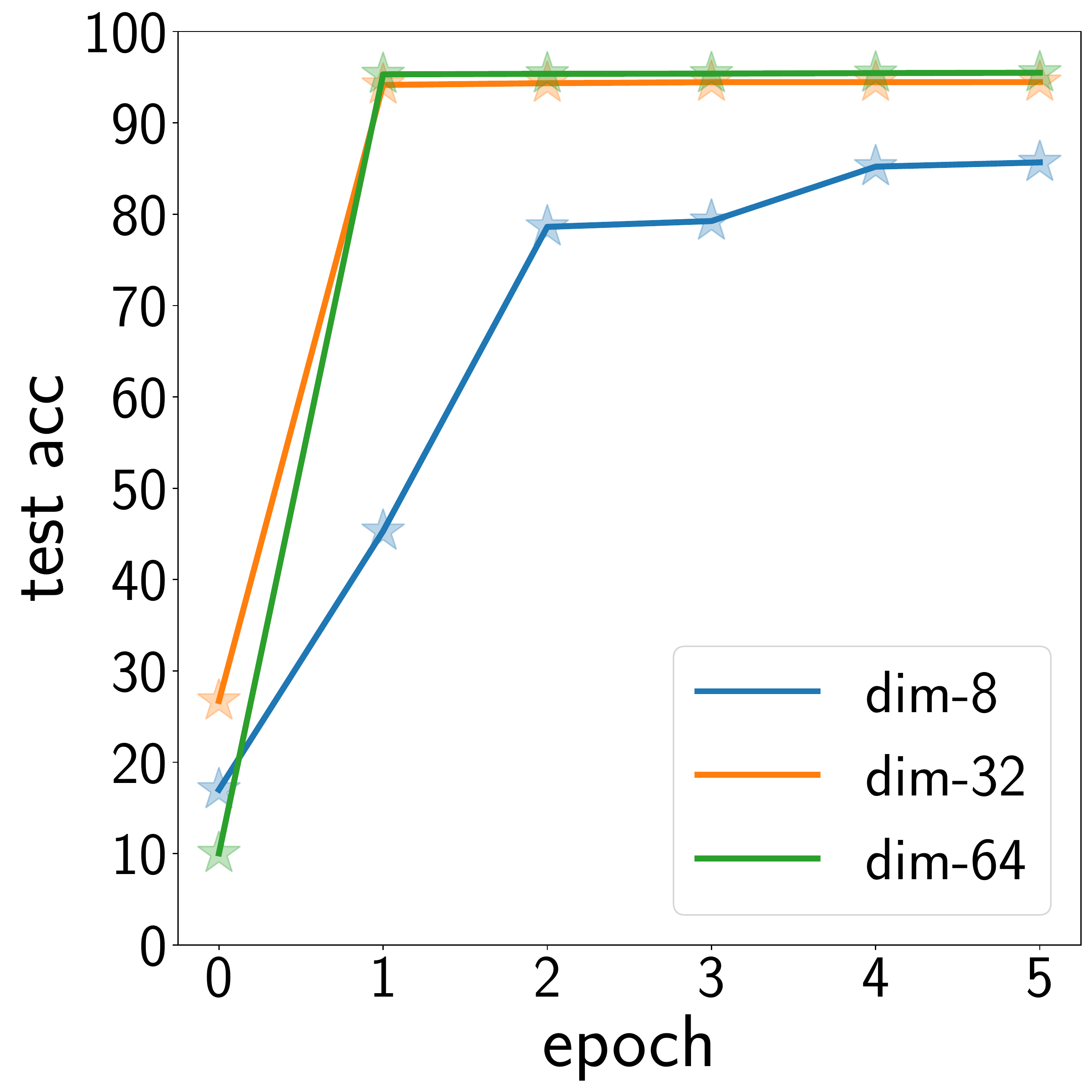}
    \caption{}
    \label{img:expcap:1}
  \end{subfigure}
  \begin{subfigure}[b]{0.31\linewidth}
    \includegraphics[width=\linewidth]{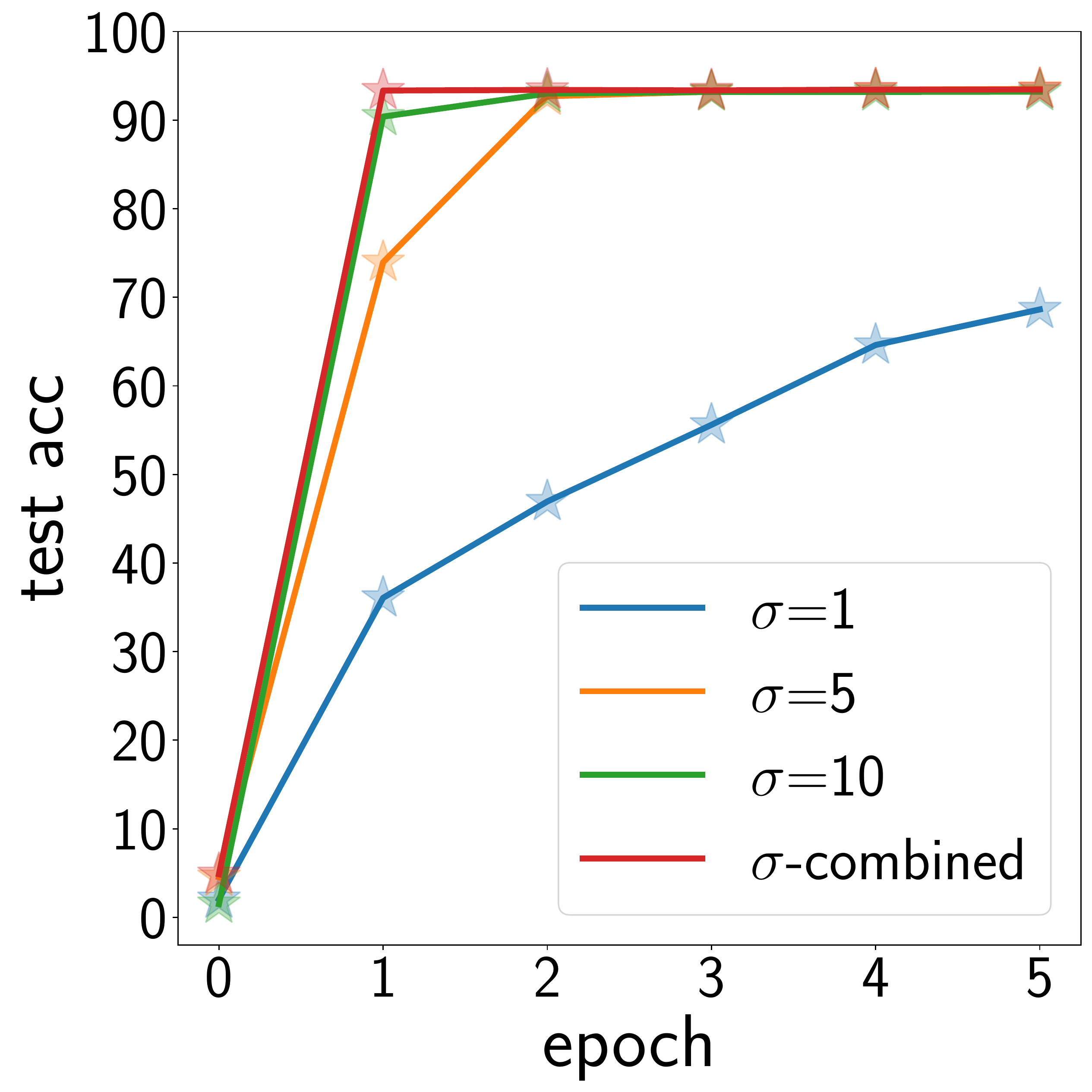}
    \caption{}
    \label{img:expcap:2}
  \end{subfigure}
  \begin{subfigure}[b]{0.31\linewidth}
    \includegraphics[width=\linewidth]{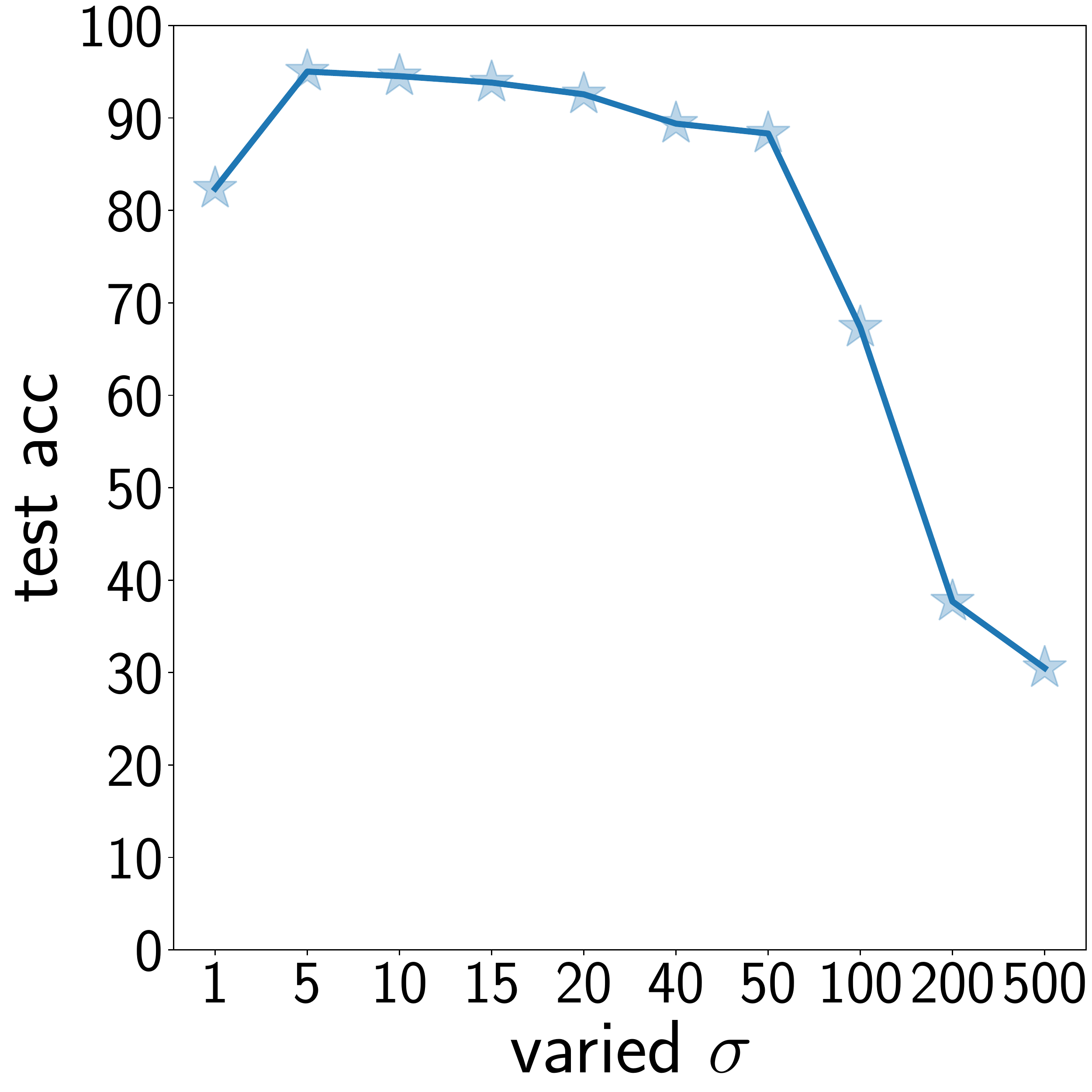}
    \caption{}
    \label{img:expcap:3}
  \end{subfigure}
  \caption{Test accuracy versus epoch number for differently sized \UNFORMATTEDED\ networks (\Fig~\ref{img:expcap:1}) and a \COMBINEDNET\  (\Fig~\ref{img:expcap:2}). The \POSTTRAINEDING\ experiment suggests that larger capacity provides more relevant information to the \POSTTRAINEDING\ stage. The \COMBINEDNET\ labelled with "$\sigma$-combined" demonstrates that using several distinct $\sigma$ values improves performance. \Fig~\ref{img:expcap:3} gives the test accuracy as a function of $\sigma$ for the network trained with one epoch.}
  \label{img:expcap}
\end{figure}

\Fig~\ref{img:expcap:1} shows the effect of different widths in the \UNFORMATTEDED\ network followed by a \POSTTRAINED\ step. The differences in the width of the \UNFORMATTEDED\ networks are reflected in the different training accuracy 
in the \POSTTRAINEDING\ stage.

\Fig~\ref{img:expcap:1} 
indicates that larger networks (say width-64 compared to width-8) lead to a faster-converging \POSTTRAINEDING\ stage. This suggests the HSIC bottleneck objective \eqref{eq:hsic01} works effectively on large networks that provide more relevant information to the \POSTTRAINEDING\ stage.  

As mentioned in previous section,  the HSIC results do depend on the chosen $\sigma$. 
In the \COMBINEDNET\ experiment (\Fig~\ref{img:expcap:2}), we aggregate several \UNFORMATTEDED\ networks with different $\sigma$ together in parallel to better capture dependencies at multiple scales. Our experiment setup trains three parallel HSIC networks having the same five-layered configuration but with different kernel widths $\sigma=1$, $\sigma=5$, and $\sigma=10$.
The resulting \POSTTRAINEDING\ performance is  shown in \Fig~\ref{img:expcap:2}. 

The results show \POSTTRAINEDING\ on the \COMBINEDNET\ outperforms other experiments, suggesting that it is providing additional information relating to the corresponding scale to the \POSTTRAINEDING\ stage. It also indicates that a single $\sigma$ is not sufficient to capture all dependencies in these networks.
Treating $\sigma$ as a learnable parameter is left for future work.

\subsection{Experiments on ResNet} \label{ss:resnet}

Our previous results aimed at demonstrating the training efficacy of the new paradigm are based on basic fully connected feedforward networks. To show that the paradigm is potentially effective for other architectures, we train a ResNet with the \UNFORMATTEDED\ framework by adding the loss \eqref{eq:hsic01} to the output of each residual block.

\begin{figure}[ht]
  \centering
  \begin{subfigure}[b]{0.32\linewidth} 
    \includegraphics[width=\linewidth]{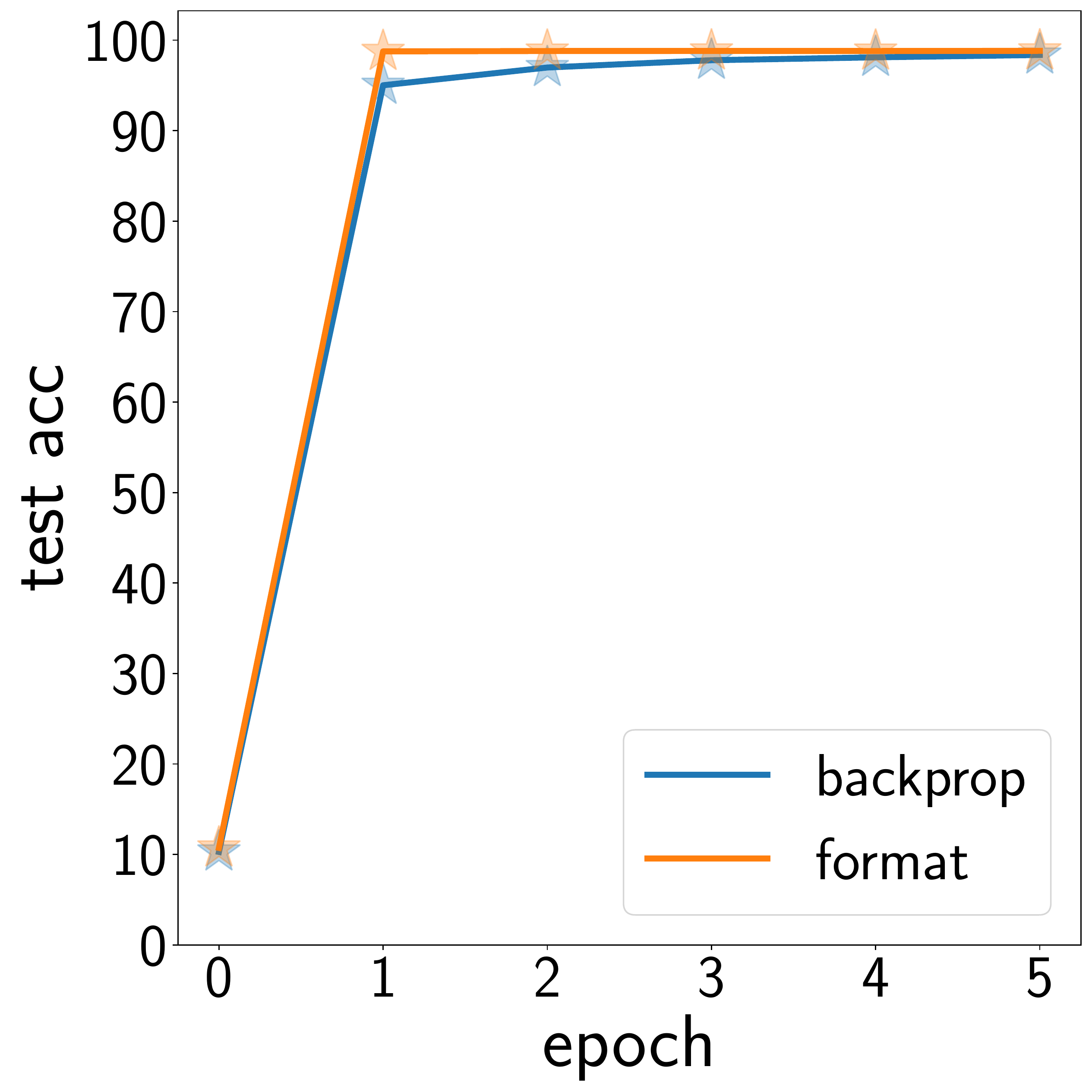}
    \caption{MNIST}
    \label{img:expres:1}
  \end{subfigure}
  \begin{subfigure}[b]{0.32\linewidth} 
    \includegraphics[width=\linewidth]{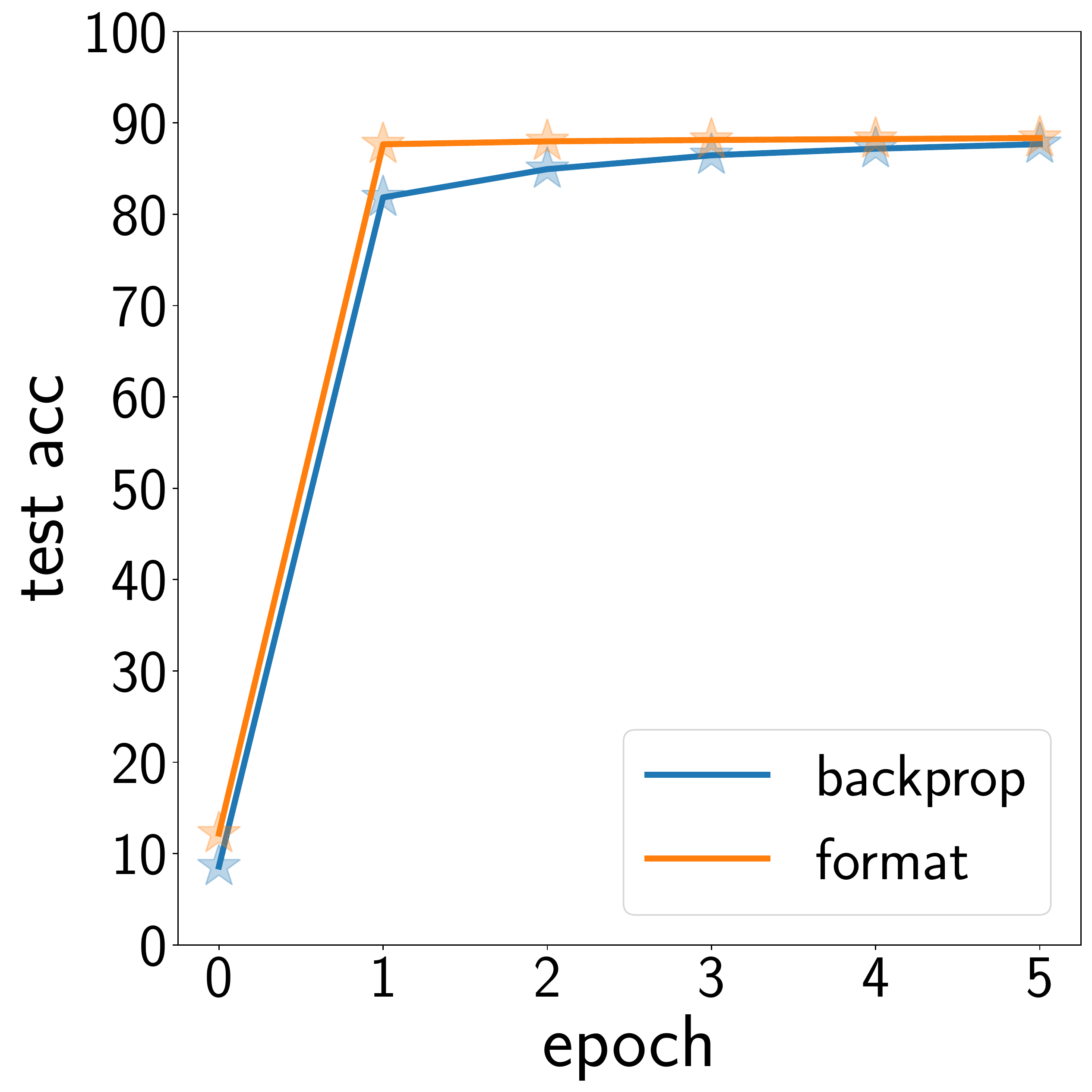}
    \caption{Fashion MNIST}
    \label{img:expres:2}
  \end{subfigure}
  \begin{subfigure}[b]{0.32\linewidth} 
    \includegraphics[width=\linewidth]{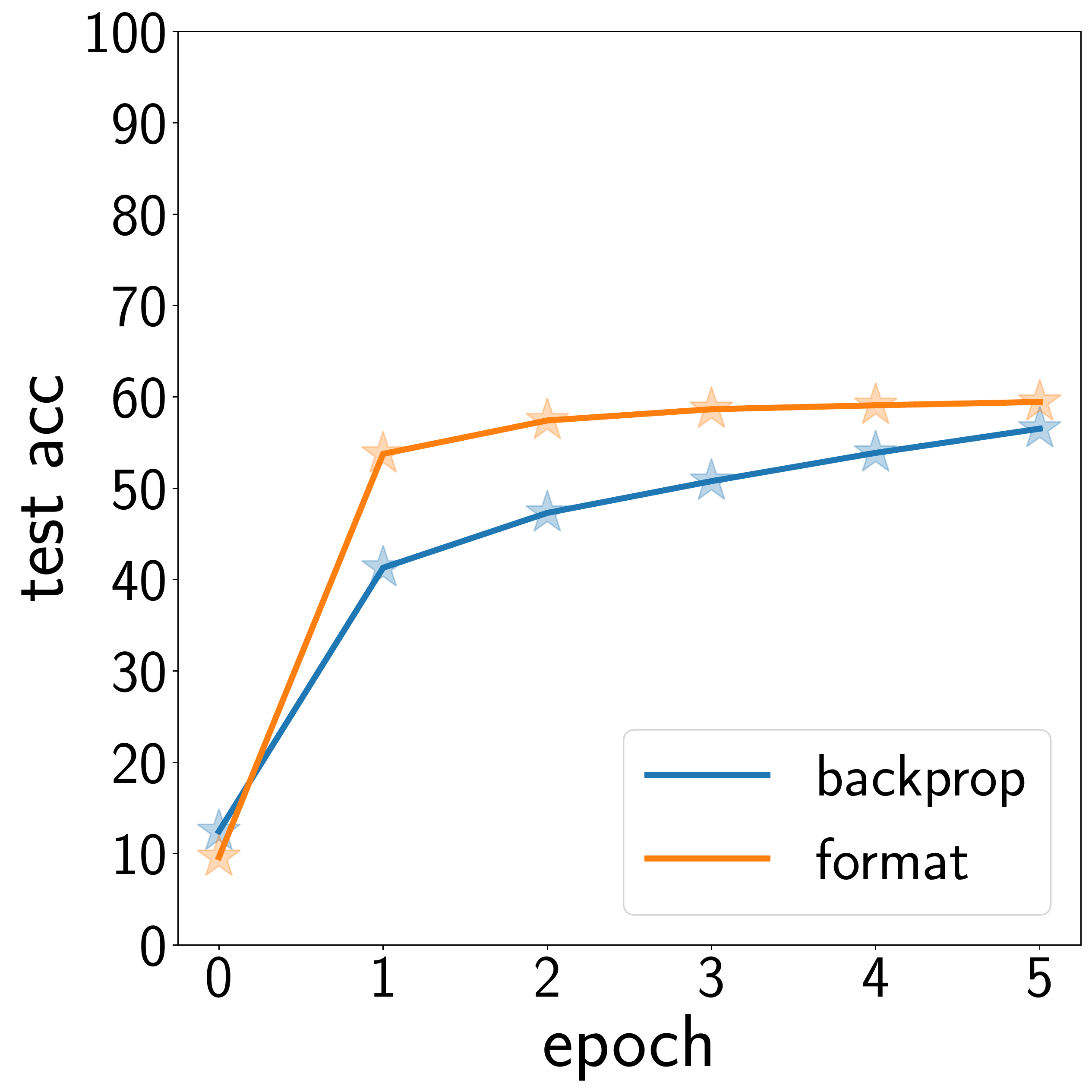}
    \caption{CIFAR10}
    \label{img:expres:3}
  \end{subfigure}
  \caption{Test accuracy versus epoch number for ResNet. The performance is shown for the comparison between the ResNet trained with \POSTTRAINEDING\ and a  backpropagation network.}
  \label{img:expres}
\end{figure}
In \Fig~\ref{img:expres}, we show the test performance for a 
network with five convolutional residual blocks on several datasets in the initial epochs. Each experiment includes five \UNFORMATTEDED\ epochs followed by \POSTTRAINEDING\ with a one-layer classifier network, and is compared with its standard backpropagation-trained counterpart. 

Our results show that the \POSTTRAINEDING\ converges more rapidly to high accuracy performance by making use of the distinct representations from the \UNFORMATTEDED\ network. 
The final test accuracy from \Fig~\ref{img:expres} is (98.8\%, 88.3\%, 59.4\%) for \POSTTRAINED\ and (98.4\%, 87.6\%, 56.5\%) for backpropagation-trained networks, for MNIST, FashionMNIST, and CIFAR10 respectively. The CIFAR10 result is well below state-of-the-art performance, because we are not using a state-of-the-art architecture. Nevertheless, the HSIC-bottleneck network provides a significant boost in convergence.

\section{Conclusion} \label{s:conclusion}

We present a new approach to train deep neural networks without the use of backpropagation. The method is inspired by the information bottleneck and can be seen as an approximation there-of,
but (to our knowledge) is the first approach that sidesteps the well known issues in computing mutual information in deep neural networks by using HSIC as a surrogate.\footnote{\cite{Vepakomma19} also introduce an HSIC-like objective for the  purpose of minimizing unnecessary dependency with the input. Their method differs in that it lacks the second term in \eqref{eq:hsic01} and cannot be interpreted as an information bottleneck scheme.}
``Unformatted'' HSIC-bottleneck training of several standard classification problems results in one-hot output that can be directly permuted to perform classification, with accuracy approximately comparable to that of standard backpropagation training of the same architectures. Performance is further improved by using the outputs as representations 
for a \POSTTRAINEDING\ stage, in which a single layer (and softmax) is appended and trained with conventional SGD, but without backpropagation. The HSIC bottleneck trained network provides good hidden representations by removing irrelevant information and retaining information that is important for the task at hand. 

HSIC bottleneck training has several benefits over conventional backpropagation:
\begin{itemize}
\item it is able to train deep networks for which backpropagation training fails (\Fig~\ref{img:presolve});
\item it mitigates the vanishing and exploding gradient issues found in conventional backpropagation, since it solves the problem layer-by-layer without the use of the chain rule;
\item it removes the need for backward sweeps;
\item it potentially allows layers to be trained in parallel, using layerwise block coordinate descent;
\item although our approach is not intended to be biologically plausible, it does address the weight transport \cite{Lillicrap2016} and update locking problems.
\end{itemize}

Our work is an initial exploration of backpropagation-free learning using the HSIC bottleneck, and, in common with other explorations of new training methods for deep learning, e.g., \cite{Choromanska19,gradienttargetprop})  does not attempt to achieve state-of-the-art performance. Future work could consider careful tuning of the $\sigma$ in HSIC to improve performance and evaluate the HSIC-bottleneck approach on different tasks such as regression and generative models.

\section*{Acknowledgments}
We thank David Balduzzi and Marcus Frean for discussions.


\small
\bibliographystyle{template/aaai}
\bibliography{reference}

\begin{thebibliography}{}

\bibitem[\protect\citeauthoryear{Alemi \bgroup et al\mbox.\egroup
  }{2017}]{alemiFischerDeep}
Alemi, A.~A.; Fischer, I.; Dillon, J.~V.; and Murphy, K.
\newblock 2017.
\newblock Deep variational information bottleneck.
\newblock In {\em ICLR}.
\newblock OpenReview.net.

\bibitem[\protect\citeauthoryear{Amjad and Geiger}{2018}]{amjadGeiger18}
Amjad, R.~A., and Geiger, B.~C.
\newblock 2018.
\newblock How (not) to train your neural network using the information
  bottleneck principle.
\newblock {\em CoRR} abs/1802.09766.

\bibitem[\protect\citeauthoryear{Baddeley, Foldiak, and
  Hancock}{1999}]{BaddeleyInfoTheoryBrain}
Baddeley, R.; Foldiak, P.; and Hancock, P., eds.
\newblock 1999.
\newblock {\em Information Theory and the Brain}.
\newblock Cambridge University Press.

\bibitem[\protect\citeauthoryear{Balduzzi, Vanchinathan, and
  Buhmann}{2015}]{balduzzi15kickback}
Balduzzi, D.; Vanchinathan, H.; and Buhmann, J.
\newblock 2015.
\newblock Kickback cuts backprop's red-tape: Biologically plausible credit
  assignment in neural networks.
\newblock In {\em Proc. AAAI}.

\bibitem[\protect\citeauthoryear{Banerjee and
  Mont{\'{u}}far}{2018}]{banerjeeDB}
Banerjee, P.~K., and Mont{\'{u}}far, G.
\newblock 2018.
\newblock The variational deficiency bottleneck.
\newblock {\em CoRR} abs/1810.11677.

\bibitem[\protect\citeauthoryear{Belghazi \bgroup et al\mbox.\egroup
  }{2018}]{Ishmael18}
Belghazi, I.; Rajeswar, S.; Baratin, A.; Hjelm, R.~D.; and Courville, A.~C.
\newblock 2018.
\newblock {MINE:} mutual information neural estimation.
\newblock {\em CoRR} abs/1801.04062.

\bibitem[\protect\citeauthoryear{Blaschko and Gretton}{2008}]{Blaschko2008b}
Blaschko, M.~B., and Gretton, A.
\newblock 2008.
\newblock A {Hilbert-Schmidt} dependence maximization approach to unsupervised
  structure discovery.
\newblock In {\em Proc.~6th Int.~Workshop on Mining and Learning with Graphs}.

\bibitem[\protect\citeauthoryear{Brakel and Bengio}{2018}]{brakel2018learning}
Brakel, P., and Bengio, Y.
\newblock 2018.
\newblock Learning independent features with adversarial nets for non-linear
  {ICA}.
\newblock \url{https://openreview.net}.

\bibitem[\protect\citeauthoryear{Choromanska \bgroup et al\mbox.\egroup
  }{2019}]{Choromanska19}
Choromanska, A.; Cowen, B.; Kumaravel, S.; Luss, R.; Rigotti, M.; Rish, I.;
  Diachille, P.; Gurev, V.; Kingsbury, B.; Tejwani, R.; and Bouneffouf, D.
\newblock 2019.
\newblock Beyond backprop: Online alternating minimization with auxiliary
  variables.
\newblock In {\em ICML},  1193--1202.

\bibitem[\protect\citeauthoryear{Cover and Thomas}{2006}]{Cover2006}
Cover, T.~M., and Thomas, J.~A.
\newblock 2006.
\newblock {\em Elements of Information Theory}.
\newblock New York, NY, USA: Wiley-Interscience.

\bibitem[\protect\citeauthoryear{de Souza~Farias and
  Maziero}{2018}]{gradienttargetprop}
de~Souza~Farias, T., and Maziero, J.
\newblock 2018.
\newblock Gradient target propagation.
\newblock {\em CoRR} abs/1810.09284.

\bibitem[\protect\citeauthoryear{Fukumizu \bgroup et al\mbox.\egroup
  }{2008}]{Fukumizu08}
Fukumizu, K.; Gretton, A.; Sun, X.; and Sch\"{o}lkopf, B.
\newblock 2008.
\newblock Kernel measures of conditional dependence.
\newblock In Platt, J.~C.; Koller, D.; Singer, Y.; and Roweis, S.~T., eds.,
  {\em NIPS}.
\newblock  489--496.

\bibitem[\protect\citeauthoryear{Goldfeld \bgroup et al\mbox.\egroup
  }{2018}]{goldfeld18}
Goldfeld, Z.; van~den Berg, E.; Greenewald, K.~H.; Melnyk, I.; Nguyen, N.;
  Kingsbury, B.; and Polyanskiy, Y.
\newblock 2018.
\newblock Estimating information flow in neural networks.
\newblock {\em CoRR} abs/1810.05728.

\bibitem[\protect\citeauthoryear{Gretton \bgroup et al\mbox.\egroup
  }{2005}]{Gretton2005}
Gretton, A.; Bousquet, O.; Smola, A.; and Sch\"{o}lkopf, B.
\newblock 2005.
\newblock Measuring statistical dependence with {H}ilbert-{S}chmidt norms.
\newblock In {\em Proc.~Int.~Conf.~Algorithmic Learning Theory},  63--77.
\newblock Springer-Verlag.

\bibitem[\protect\citeauthoryear{Ioffe and Szegedy}{2015}]{IoffeS15}
Ioffe, S., and Szegedy, C.
\newblock 2015.
\newblock Batch normalization: Accelerating deep network training by reducing
  internal covariate shift.
\newblock {\em CoRR} abs/1502.03167.

\bibitem[\protect\citeauthoryear{Kohan, Rietman, and
  Siegelmann}{2018}]{KohanErrorforward18}
Kohan, A.~A.; Rietman, E.~A.; and Siegelmann, H.~T.
\newblock 2018.
\newblock Error forward-propagation: Reusing feedforward connections to
  propagate errors in deep learning.
\newblock {\em CoRR} abs/1808.03357.

\bibitem[\protect\citeauthoryear{Kolchinsky, Tracey, and
  Wolpert}{2017}]{kolchinskyNIB}
Kolchinsky, A.; Tracey, B.~D.; and Wolpert, D.~H.
\newblock 2017.
\newblock Nonlinear information bottleneck.
\newblock {\em CoRR} abs/1705.02436.

\bibitem[\protect\citeauthoryear{{Kwak} and {Chong-Ho Choi}}{2002}]{Kwak02}
{Kwak}, N., and {Chong-Ho Choi}.
\newblock 2002.
\newblock Input feature selection by mutual information based on {P}arzen
  window.
\newblock {\em IEEE Tran.~Pattern Analysis and Machine Intelligence}
  24(12):1667--1671.

\bibitem[\protect\citeauthoryear{Lillicrap \bgroup et al\mbox.\egroup
  }{2016}]{Lillicrap2016}
Lillicrap, T.~P.; Cownden, D.; Tweed, D.~B.; and Akerman, C.~J.
\newblock 2016.
\newblock Random synaptic feedback weights support error backpropagation for
  deep learning.
\newblock {\em Nature Communications} 7.

\bibitem[\protect\citeauthoryear{Lopez \bgroup et al\mbox.\egroup
  }{2018}]{Romain18}
Lopez, R.; Regier, J.; Yosef, N.; and Jordan, M.~I.
\newblock 2018.
\newblock Information constraints on auto-encoding variational bayes.
\newblock {\em CoRR} abs/1805.08672.

\bibitem[\protect\citeauthoryear{Moskovitz, Litwin{-}Kumar, and
  Abbott}{2018}]{Theodore2018}
Moskovitz, T.~H.; Litwin{-}Kumar, A.; and Abbott, L.~F.
\newblock 2018.
\newblock Feedback alignment in deep convolutional networks.
\newblock {\em CoRR} abs/1812.06488.

\bibitem[\protect\citeauthoryear{Saxe \bgroup et al\mbox.\egroup
  }{2018}]{michael2018on}
Saxe, A.~M.; Bansal, Y.; Dapello, J.; Advani, M.; Kolchinsky, A.; Tracey,
  B.~D.; and Cox, D.~D.
\newblock 2018.
\newblock On the information bottleneck theory of deep learning.
\newblock In {\em International Conference on Learning Representations}.

\bibitem[\protect\citeauthoryear{Sejdinovic \bgroup et al\mbox.\egroup
  }{2012}]{Dino12}
Sejdinovic, D.; Gretton, A.; Sriperumbudur, B.~K.; and Fukumizu, K.
\newblock 2012.
\newblock Hypothesis testing using pairwise distances and associated kernels
  (with appendix).
\newblock {\em CoRR} abs/1205.0411.

\bibitem[\protect\citeauthoryear{Shwartz{-}Ziv and
  Tishby}{2017}]{ShwartzZivT17}
Shwartz{-}Ziv, R., and Tishby, N.
\newblock 2017.
\newblock Opening the black box of deep neural networks via information.
\newblock {\em CoRR} abs/1703.00810.

\bibitem[\protect\citeauthoryear{Sriperumbudur, Fukumizu, and
  Lanckriet}{2010}]{sriperumbudur2010relation}
Sriperumbudur, B.~K.; Fukumizu, K.; and Lanckriet, G.
\newblock 2010.
\newblock On the relation between universality, characteristic kernels and
  {RKHS} embedding of measures.
\newblock In {\em International Conference on Artificial Intelligence and
  Statistics},  773--780.

\bibitem[\protect\citeauthoryear{Sugiyama and Yamada}{2012}]{Sugiyama12}
Sugiyama, M., and Yamada, M.
\newblock 2012.
\newblock On kernel parameter selection in {H}ilbert-{S}chmidt independence
  criterion.
\newblock {\em IEICE Transactions on Information and Systems} E95D.

\bibitem[\protect\citeauthoryear{Tishby, Pereira, and
  Bialek}{1999}]{Tishby99theinformation}
Tishby, N.; Pereira, F.~C.; and Bialek, W.
\newblock 1999.
\newblock The information bottleneck method.
\newblock In {\em Allerton Conference on Communication, Control, and
  Computing}.

\bibitem[\protect\citeauthoryear{Tschannen \bgroup et al\mbox.\egroup
  }{2019}]{tschannenMI19}
Tschannen, M.; Djolonga, J.; Rubenstein, P.~K.; Gelly, S.; and Lucic, M.
\newblock 2019.
\newblock On mutual information maximization for representation learning.
\newblock {\em CoRR} abs/1907.13625.

\bibitem[\protect\citeauthoryear{Vepakomma \bgroup et al\mbox.\egroup
  }{2019}]{Vepakomma19}
Vepakomma, P.; Gupta, O.; Dubey, A.; and Raskar, R.
\newblock 2019.
\newblock Reducing leakage in distributed deep learning for sensitive health
  data.
\newblock In {\em {ICLR} {AI} for social good workshop}.

\bibitem[\protect\citeauthoryear{{Werbos}}{1990}]{Werbos90}
{Werbos}, P.~J.
\newblock 1990.
\newblock Backpropagation through time: what it does and how to do it.
\newblock {\em Proceedings of the IEEE} 78(10):1550--1560.

\bibitem[\protect\citeauthoryear{Wu \bgroup et al\mbox.\egroup }{2018}]{Wu2018}
Wu, D.; Zhao, Y.; Tsai, Y.-H.~H.; Yamada, M.; and Salakhutdinov, R.
\newblock 2018.
\newblock "{D}ependency {B}ottleneck" in auto-encoding architectures: an
  empirical study.
\newblock {\em CoRR} abs/1802.05408.

\end{thebibliography}

\twocolumn[\section*{The HSIC Bottleneck: Supplementary Material\\~\ \\~\ }] \label{s:spp}  \label{s:spp}

\subsection*{Relating HSIC to Entropy}

Although HSIC is a measure of (in)dependence, its exact relation to mutual information has not been established. Mutual information is defined in terms of entropy, and it has been observed that entropy and Fisher information are related as volume and surface area \cite{Cover2006}. 

In this section we outline an informal argument that HSIC is analogously related to diameter,
making use of the identity $I(X,X) = H(X)$.  
The relation of entropy to volume is easily suggested in the case of a multivariate Gaussian RV with covariance matrix $\CC$.  
The entropy in this case is 
\begin{align*}
   H(X_{\text{gaussian}}) &= \frac12 \log \left( (2\pi e)^d |\CC| \right)
\\
	&\propto \log \det \CC \quad+\quad\text{constant} 
\end{align*}
Here the interpretation as a ``volume'' can be seen through the presence of the determinant.
For comparison with HSIC (below), note that $\CC$ is symmetric and can be diagonalized with real eigenvalues,
\[
	\det \CC = \det \UU \LLambda \UU^T = \det \LLambda = \prod \lambda_k
\]

Turning to HSIC, denoting $\tilde{\KK}_X = \HH \KK_X \HH$ (and similarly for $\YY$, 
the HSIC computation is $\frac{1}{n^2} \tr \tilde{\KK}_X \tilde{\KK}_Y$. 
HSIC(X,X) is thus simply $\frac{1}{n^2} \tr \tilde{\KK}_X^2$, i.e. the (scaled) squared Frobenious norm of the feature covariance matrix.

Now in general $\| \bAA \|_F^2 = \sum \lambda_k^2$ because
\[
   \tr \bAA^T \bAA = \tr (\PP \LLambda \PP^{-1})( \PP \LLambda \PP^{-1}) = \tr \PP \LLambda^2 \PP^{-1} = \tr \LLambda^2
\]
making use of the Jordan normal decomposition
(or eigendecomposition in our case because $\tilde{\KK}_X$ is symmetric).

In summary, entropy is related to volume, as can be see in the Gaussian case through the presence of the determinant, i.e. the product of the eigenvalues.  HSIC(X,X) is related to the Frobenius norm, which is a \emph{sum} of eigenvalues, i.e.~a ``diameter''.

\subsection*{Toy \UNFORMATTED\ solve}

\begin{figure}[H]
  \centering
  \begin{subfigure}[b]{0.45\linewidth} \label{fig_1}
    \includegraphics[width=\linewidth]{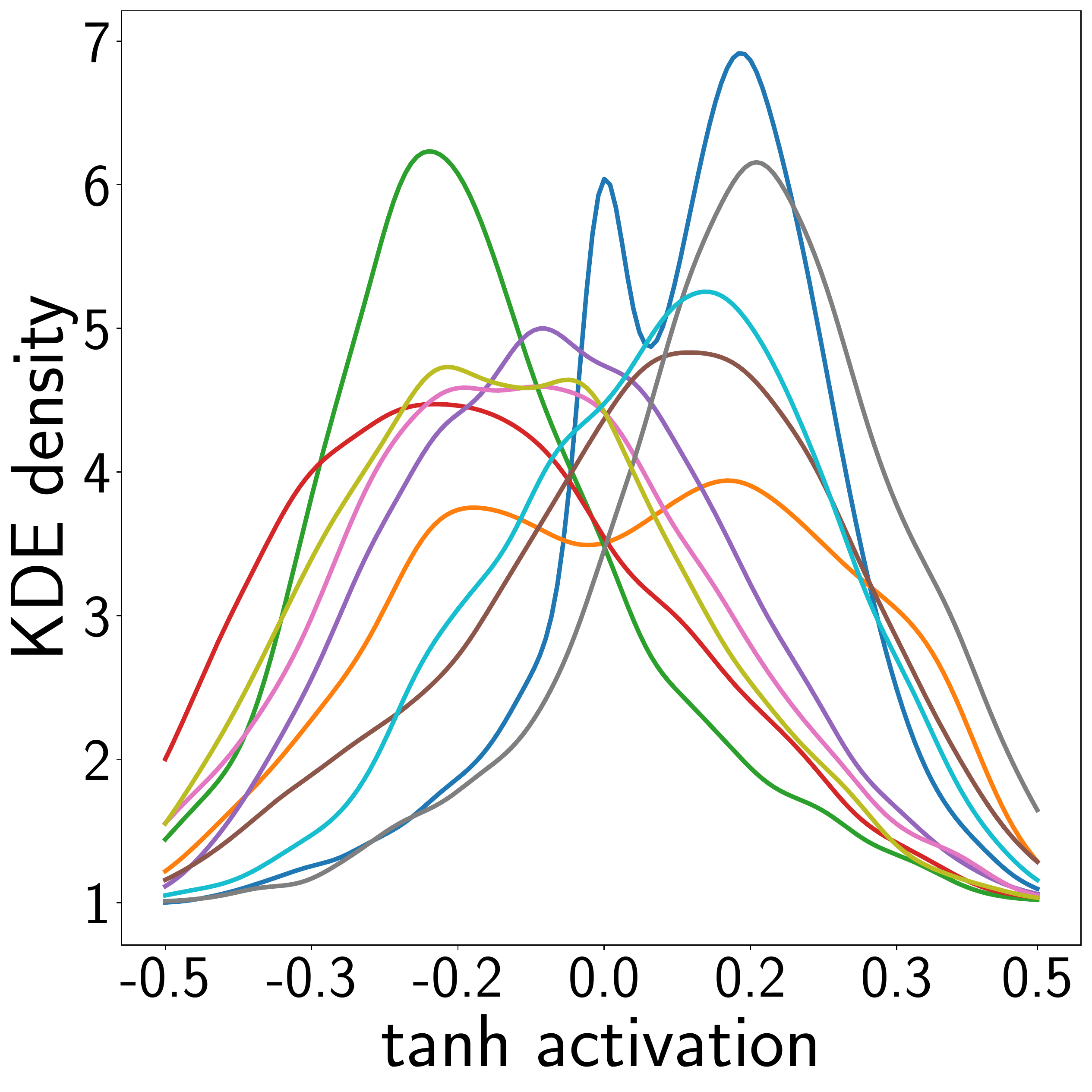}
    \caption{Backpropagation}
    \label{img:needle:a}
  \end{subfigure}
  \hspace{1mm}
  \begin{subfigure}[b]{0.45\linewidth} \label{fig_2}
    \includegraphics[width=\linewidth]{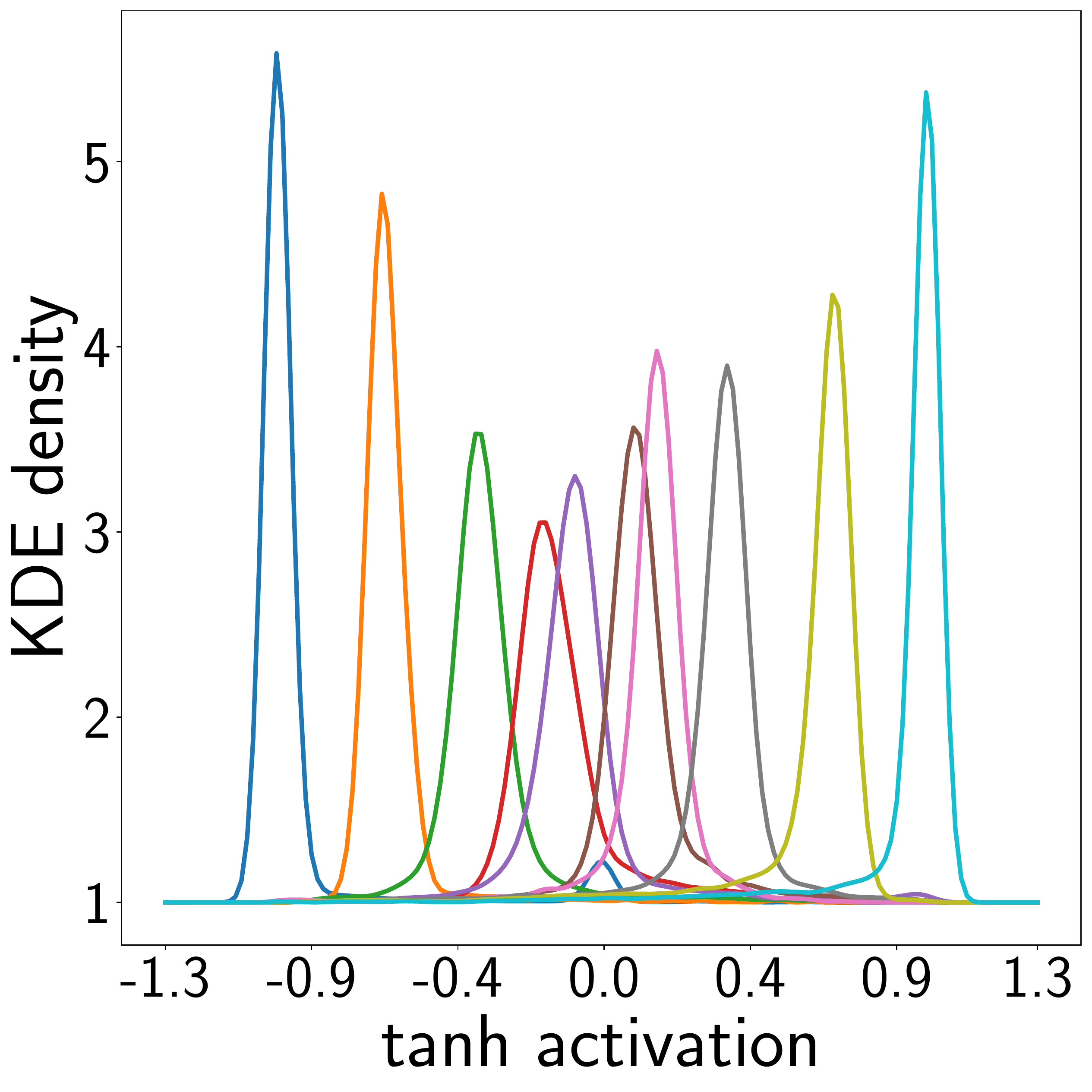}
    \caption{\UNFORMATTEDC}
    \label{img:needle:b}
  \end{subfigure}
  \caption{The tanh classed activation in $\Reals^{1}$ distribution of MNIST between backpropagation Fig~\ref{img:needle:a}, and \UNFORMATTED\ Fig~\ref{img:needle:b}. Each curve represents the activation distribution from particular image category. The distribution is generated by Gaussian Kernel Density Estimation (KDE).}
  \label{img:needle}
\end{figure}

\Fig~\ref{img:presolveonehot} has shown that \UNFORMATTED\ discriminates class signals during the training, resulting a one-hot effect. We did a further experiment based on a simple model sized 784-256-128-64-32-16-8-\textbf{1}  with backpropagation and \UNFORMATTED. As is convention, we appended an output layer $O \in \Reals^{10}$ to the backpropagation model to support the cross-entropy objective. 
The goal of this experiment is to see how well the \UNFORMATTED\ can separate the signal classes in $\Reals^1$.  Although the results change based on the random seed, we found that \UNFORMATTED\ usually produces better separation than backpropagation (\Fig~\ref{img:needle}).



\subsection*{Differences in Arxiv version 2}
All the experimental results are new and generally show the test performance over additional epochs. The figures are improved to be more legible. The experiment hyperparameters benefit from a modest hyperparameter search, whereas the hyperparameters in the first draft were manually chosen and did not reflect the best performance of either our method or the baseline. 
The text has been clarified to emphasize that we do not claim biological plausibiltiy; rather, our method merely shares characteristics (no backpropagation, weight transport, or update locking) with other recent methods that are regarded as making some progress toward this goal.


\end{document}